\documentclass{article}

\usepackage{microtype}
\usepackage{graphicx}
\usepackage{subcaption}
\usepackage{booktabs} 

\usepackage{hyperref}

\usepackage[preprint]{icml2026}

\usepackage{amsmath}
\usepackage{amssymb}
\usepackage{mathtools}
\usepackage{amsthm}

\usepackage{hyperref}
\usepackage{url}
\usepackage{graphicx}
\usepackage{subcaption}
\usepackage{booktabs}
\usepackage{array}
\usepackage{multirow}
\usepackage{xcolor}
\usepackage{colortbl}
\usepackage{enumitem}
\usepackage[dvipsnames]{xcolor}
\usepackage{wrapfig}
\usepackage{textcomp}
\usepackage{caption}

\usepackage[capitalize,noabbrev]{cleveref}

\theoremstyle{plain}

\theoremstyle{definition}

\theoremstyle{remark}

\usepackage[textsize=tiny]{todonotes}

\icmltitlerunning{Degraded Image Restoration via Latent Rectified Flow \& Feature Distillation}

\begin{document}

\twocolumn[
  \icmltitle{Degraded Image Restoration via Latent Rectified Flow \& Feature Distillation}



  \icmlsetsymbol{equal}{*}

  \begin{icmlauthorlist}
    \icmlauthor{Shourya Verma}{equal,1}
    \icmlauthor{Mengbo Wang}{equal,1}
    \icmlauthor{Ananth Grama}{1}
    \icmlauthor{Nadia Atallah Lanman}{2}
  \end{icmlauthorlist}

  \icmlaffiliation{1}{Department of Computer Science, Purdue University, USA}
  \icmlaffiliation{2}{Purdue Institute of Cancer Research, USA}

  \icmlcorrespondingauthor{Shourya Verma}{verma198@purdue.edu}

  \icmlkeywords{Machine Learning, ICML}

  \vskip 0.3in
]



\printAffiliationsAndNotice{}  

\begin{abstract}
Current approaches for restoration of degraded images face a trade-off: high-performance models are slow for practical use, while fast models produce poor results. Knowledge distillation transfers teacher knowledge to students, but existing static feature matching methods cannot capture how modern transformer architectures dynamically generate features. We propose a novel Latent Rectified Flow Feature Distillation method for restoring degraded images called \textbf{'RestoRect'}. We apply rectified flow to reformulate feature distillation as a generative process where students learn to synthesize teacher-quality features through learnable trajectories in latent space. Our framework combines Retinex decomposition with learnable anisotropic diffusion constraints, and trigonometric color space polarization. We introduce a Feature Layer Extraction loss for robust knowledge transfer between different network architectures through cross-normalized transformer feature alignment with percentile-based outlier detection. RestoRect achieves better training stability, and faster convergence and inference while preserving restoration quality, demonstrating superior results across 15 image restoration datasets, covering 4 tasks, on 10 metrics against baselines. 
\end{abstract}

\section{Introduction}

\begin{figure*}[h]
    \centering
    \includegraphics[width=1\linewidth]{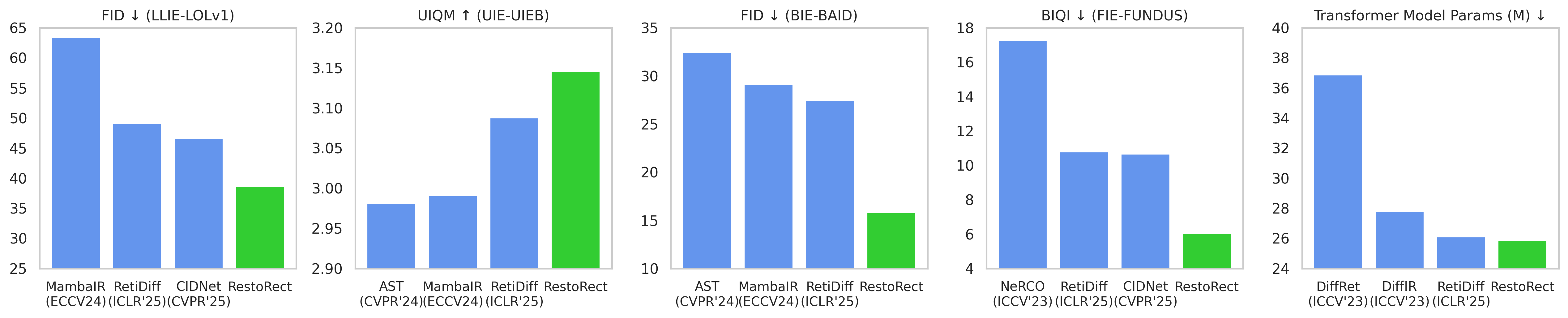}
    \caption{RestoRect achieves superior performance on four image restoration tasks while keeping parameter count low}
    \label{fig:main}
\end{figure*}

Image restoration from degraded inputs including low-light (LLIE), underwater (UIE), backlit (BAID), and fundus (FIE) enhancement remains a key challenge in computer vision. Real-world images suffer from illumination degradation, noise, and compression artifacts that impair perception and downstream tasks. Traditional optimization methods exploit physical priors but fail on complex degradations. Transformer-based deep learning achieves strong restoration through multi-scale features, while diffusion models operating in latent spaces with Retinex priors capture complex natural image distributions. However, diffusion methods incur steep computational costs limiting real-time use. Knowledge distillation transfers knowledge from large teachers to compact students but struggles with transformer-based restoration. Conventional approaches compute static feature losses between layers, neglecting dynamic feature generation of multi-head attention and layer interactions. This mismatch hampers dependency modeling and degrades student performance. Recent models such as Reti-Diff \citep{he2023reti} and \hyperref[ac]{HVI}-CIDNet \citep{yan2024you} achieve strong restoration but rely on static feature matching that fails to capture generative processes.

We propose \textbf{RestoRect}, which formulates knowledge distillation as a generative process through latent rectified flow. Student networks learn dynamic feature synthesis via flow matching dynamics using linear interpolation trajectories between noise and target features in latent space, reducing sampling steps while preserving quality. The core \textbf{F}eature \textbf{L}ayer \textbf{EX}traction (FLEX) Loss addresses distribution mismatch by normalizing teacher and student features using student statistics, enabling meaningful comparison despite evolving distributions. Percentile-based outlier detection mitigates noisy regions. The framework integrates Retinex illumination-decomposition, learnable anisotropic diffusion for structural consistency, and trigonometric color space polarization to eliminate red discontinuity artifacts. RestoRect employs two-stage training. Stage 1 trains the teacher with pixel and perceptual losses to achieve high-quality restoration. Stage 2 distills knowledge via latent rectified flow in two phases. Phase 1 trains only rectified flow velocity predictors while freezing the restoration network. The teacher extracts Retinex and image features from degraded and ground-truth pairs as targets for two rectified flow models that learn velocity fields reproducing teacher features through learnable trajectories, enabling synthesis in only a few steps. Phase 2 trains the full restoration network: velocity predictors generate student features aligned with teacher features via \hyperref[ac]{FLEX} Loss that cross-normalizes multi-scale transformer representations and applies percentile-based outlier detection. This enables the student to efficiently generate teacher-quality features, achieving diffusion-level restoration performance at significantly higher efficiency.

Our key technical contributions include: \textbf{1.} A novel framework modeling knowledge transfer as a generative process using latent rectified flow, where the student network learns velocity fields to synthesize teacher-quality features. \textbf{2.} A novel U-Net transformer architecture with Spatial Channel Layer Normalization (SCLN) and Query-Key normalization, for attention stability under degraded inputs. \textbf{3.} A novel Feature Layer EXtraction (FLEX) Loss using feature statistics to normalize both teacher and student representations for multi-scale alignment in transformers. \textbf{4.} Combining known Retinex theory with learnable anisotropic diffusion constraints and trigonometric color space polarization to eliminate artifacts and boost restoration quality.

\section{Related Work}

\textbf{Degraded Image Restoration} has evolved from classical signal processing to modern deep learning frameworks. Early approaches such as histogram equalization~\citep{cheng2004simple}, gamma correction~\citep{huang2012efficient}, and Retinex theory~\citep{edwin1977retinex} provided interpretable solutions but failed to generalize across degradations. Retinex-based extensions~\citep{fu2016weighted,li2018structure} incorporated physical priors for reflectance–illumination decomposition, yet remained constrained by hand-crafted assumptions. Deep learning enabled data-driven feature learning, with convolutional models by ~\citep{wei2018deep} and by ~\citep{wang2019underexposed} leveraging Retinex decomposition for improved color correction. Transformer-based methods further enhanced global illumination consistency~\citep{zamir2022learning}, while adaptive designs by ~\citep{xu2022snr} and state space models like by ~\citep{guo2024mambair} advanced efficiency and context modeling. Specialized solutions addressed low-light enhancement~\citep{guo2020zero,jiang2021enlightengan}, underwater restoration~\citep{naik2021shallow,guo2023underwater}, and backlit enhancement~\citep{gaintseva2024rave,jiang2021enlightengan}. Hybrid approaches such as by ~\citep{he2025unfoldir} bridged optimization- and learning-based paradigms via deep unfolding, while by ~\citep{yan2024you,yan2025hvi} introduced learnable color-space transformations to decouple brightness and chromaticity.

\textbf{Image Generative Modeling} aims to capture complex data distributions and synthesize realistic details. GAN-based methods ~\citep{cong2023pugan,jiang2021enlightengan} achieved high-quality results but suffered from instability and mode collapse. Diffusion models improved fidelity through iterative denoising~\citep{yi2023diff}, though efficiency remained limited. Latent-space diffusion, such as Reti-Diff~\citep{he2023reti}, reduced overhead by incorporating Retinex priors. Flow-based approaches offered exact likelihoods and stable training~\citep{kingma2018glow}, with rectified flow~\citep{liu2022flow} enabling efficient straight-line sampling. Integrating generative priors into restoration networks has driven advances in knowledge distillation~\citep{hinton2015distilling}, conditional and multi-scale generation~\citep{saharia2022palette,ho2022imagen}, and physics-informed restoration~\citep{xia2023diffir}. Nonetheless, achieving real-time, high-fidelity restoration remains challenging due to the trade-off between generative quality and computational efficiency.

\textbf{Knowledge Distillation} enables compact models to inherit capabilities from larger teachers~\citep{hinton2015distilling}. Early methods matched intermediate features~\citep{romero2014fitnets} or attention maps~\citep{zagoruyko2016paying}, using L2 losses~\citep{heo2019comprehensive} or attention transfer~\citep{huang2017like}. For vision transformers, challenges from multi-head attention and positional encodings inspired approaches like distillation tokens in DeiT~\citep{touvron2021training} and attention matrix alignment~\citep{wang2020minilm}. However, these strategies treat features as static targets, overlooking the dynamic generation in transformer architectures~\citep{jiao2019tinybert}. In image restoration, distillation is further complicated by multi-scale feature dependencies and complex distributions~\citep{zhang2022nested,berrada2025boosting}. Architectural mismatches between teacher and student amplify these gaps, limiting transfer efficiency and degrading restoration quality, motivating new paradigms that model feature generation as a learnable process rather than static matching \citep{bing2025optimizing}.

\section{Methodology}

We tackle efficient knowledge distillation for degraded image restoration, aiming to transfer knowledge from a powerful teacher $\mathcal{F}_T$ to a lightweight student $\mathcal{F}_S$ without sacrificing quality. Given a degraded input $I_{LQ} \in \mathbb{R}^{H \times W \times 3}$ and ground truth $I_{GT} \in \mathbb{R}^{H \times W \times 3}$, the objective is: $\mathcal{F}_S(I_{LQ}) \approx \mathcal{F}_T(I_{LQ}) \approx I_{GT}$. The main challenge is feature distribution mismatch between teacher and student. Standard distillation aligns features with simple distance metrics, which breaks down when distributions differ significantly, especially in transformer-based networks where multi-head attention produces features with varying means, variances, and outlier characteristics.

\begin{figure*}[htbp]
    \centering
    \includegraphics[width=1\linewidth]{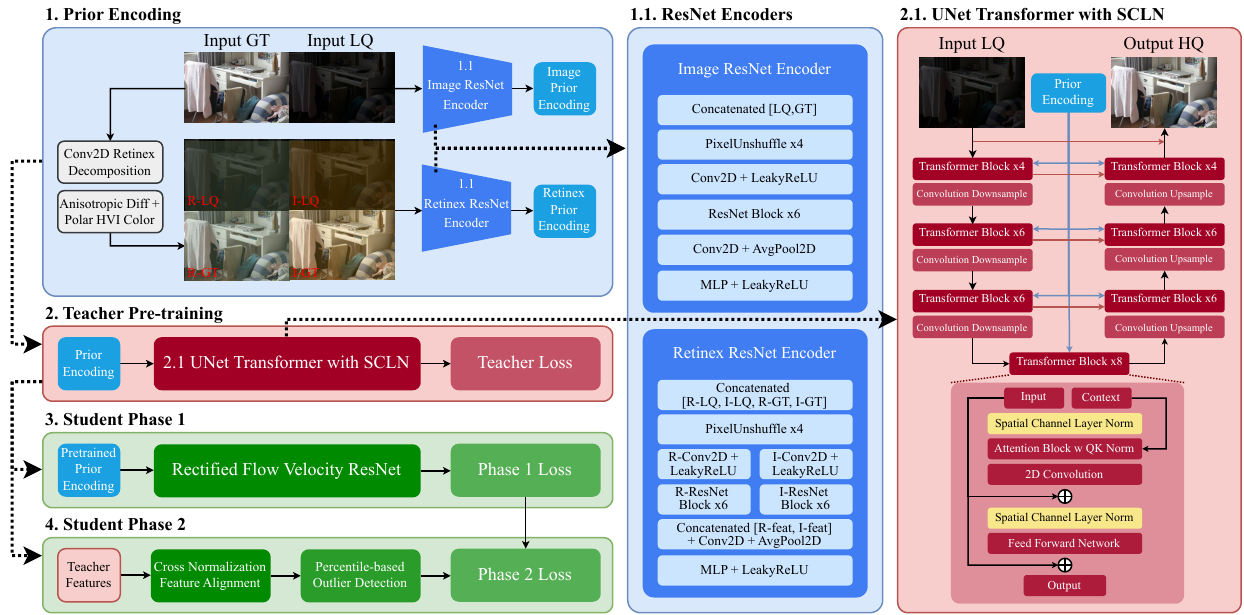}
    \caption{Training framework flowchart for RestoRect. Starting from top left (1. Prior Encoding) the inputs go through retinex decomposition and pass through encoders (1.1 ResNet Encoders) to prepare image and retinex prior encodings. Next these prior encodings are pre-trained (2. Teacher Pre-training) with the teacher model (2.1 UNet Transformer with SCLN) using a reconstruction loss. Finally the frozen prior encodings and teacher model are used for student phase 1 and phase 2 training using rectified flow loss. Full architecture details in Appendix \ref{sec:arch}.}
    \label{fig:arch}
\end{figure*}

\subsection{Teacher Network Transformer Pretraining}

Our method uses well-established Retinex theory to derive illumination-informed features as priors for knowledge distillation. Retinex models an image $I$ as the product of reflectance $R$ and illumination $L$: $I = R \odot L$, where $R$ encodes surface properties and $L$ captures lighting. We use two decomposition networks, $\mathcal{D}_l$ (low-light) and $\mathcal{D}_h$ (normal-light), each mapping $\mathcal{D}(I) \rightarrow (R, L)$ with $R \in \mathbb{R}^{H \times W \times 3}$ and $L \in \mathbb{R}^{H \times W \times 1}$ \citep{wu2022uretinex,he2023reti}. This dual setup ensures robust decomposition under diverse lighting. The decomposed components are then encoded (Figure \ref{fig:arch}(1)): a Retinex encoder extracts features from $[R;L]$ via reflectance (192-dim) and illumination (64-dim) pathways, while an image encoder processes raw image features to preserve holistic appearance. Our teacher network uses U-Net transformer architecture \citep{huang2020unet, cao2022swin} with key innovations for robust image restoration. The hierarchical transformer architecture processes multi-scale representations through encoder-decoder structures with skip connections, incorporating specialized normalization and attention mechanisms designed for degraded image inputs. Traditional layer normalization operates independently on spatial and channel dimensions, potentially losing critical spatial correlations essential for restoration tasks.

\textbf{Spatial Channel Layer Normalization (SCLN)} is introduced that captures global image statistics: $\text{SCLN}(x) = (x - \mu_{global})/(\sqrt{\sigma_{global}^2 + \epsilon}) \cdot \gamma$, where the global statistics are computed across flattened spatial-channel dimensions. This novel formulation ensures that normalization captures both local spatial patterns and global image characteristics, with learnable channel-wise scaling $\gamma \in \mathbb{R}^C$ that adapts to different feature semantics. Transformer-based restoration suffers from attention instability during training, particularly with degraded inputs which have irregular noise patterns and missing information. We apply normalization to query and key representations before attention computation, which prevents attention weight saturation in degraded regions, and ensures stable gradients throughout the attention mechanism: $\text{Attn}(Q, K, V) = \text{softmax}\left(\frac{\text{Norm}(Q) \cdot \text{Norm}(K)^T}{\sqrt{d_k}} \cdot \tau\right) V$. The teacher network processes both raw images and their Retinex decompositions through separate pathways. This design allows queries from reflectance components to attend to illumination structure, preserving intrinsic scene properties.

The benchmark results in Figure \ref{fig:norm} presents a comprehensive comparison of normalization strategies for image restoration transformers. We evaluate SCLN against LayerNorm (per-token, the baseline used in Reti-Diff), RMSNorm (per-token, no mean subtraction), and i-LN (\cite{lee2025analyzing}, norm with global-std rescaling). Row 1 visualizes channel-0 feature maps after normalization of a synthetic degraded input with channel imbalance and outlier pixels. Per-token methods (RMSNorm, LayerNorm) destroy the spatial structure visible in the input, while SCLN preserves it. i-LN preserves spatial correlations but the global-std rescaling saturates activations into extreme values, visible as high-contrast noise. This is confirmed quantitatively: the spatial autocorrelation plot shows SCLN and i-LN both match the input correlation (0.268), while per-token methods collapse to near zero. However, i-LN's output distribution exhibits heavy tails extending beyond +/-4, whereas SCLN produces the tightest distribution centered at zero, which is ideal for stable downstream attention. The per-channel std plot reveals the mechanism: i-LN reintroduces the original magnitude scale, producing std spikes an order of magnitude above other methods on high-variance channels. Training dynamics (rightmost column) track per-channel squared mean across a 6-layer transformer stack, following the protocol of \cite{lee2025analyzing} (Fig. 1-2). i-LN features diverge worst (reaching 80+ at deeper layers), as the rescaling compounds through residual connections. LayerNorm diverges moderately as the network learns to bypass per-token normalization to recover spatial structure. SCLN remains flat throughout, as normalization preserves spatial structure without rescaling, eliminating the bypass incentive entirely. SCLN achieves this at the same latency as LayerNorm (0.70ms), while i-LN incurs 20\% overhead from the rescaling computation.

\textbf{Auxiliary Constraints} like anisotropic diffusion \citep{perona1994anisotropic} and polarized \hyperref[ac]{HVI} color spaces \citep{yan2024you} \citep{yan2025hvi} are incorporated that enforce edge-preserving texture matching and eliminate artifacts. The anisotropic diffusion operator computes: $\mathcal{A}(I) = \nabla \cdot (c(|\nabla I|) \nabla I)$, with the diffusion coefficient defined as: $c(|\nabla I|) = \exp\left(-|\nabla I|^2/s^2\right)$, where $s$ is a learnable sensitivity parameter initialized as $s = 0.1$ and constrained to $s \in [0.01, 1.0]$ to prevent numerical instability. The texture consistency loss enforces structural similarity between input and predicted reflectance: $L_{tex} = \|\mathcal{A}(I_{input}) - \mathcal{A}(R_{pred})\|_1$. This constraint preserves essential edge structures while suppressing noise, maintains texture coherence across different scales, and provides gradient-based supervision for fine-grained details. 

Standard image color spaces exhibit critical limitations for restoration like discontinuities at the red boundary ($H = 0^{\circ}$ and $H = 360^{\circ}$) and degenerate mappings in dark regions. To address these fundamental limitations, polarized \hyperref[ac]{HVI} (Horizontal-Vertical-Intensity) color space is introduced that eliminates these artifacts through trigonometric parameterization. The polarized transformation maps hue to continuous coordinates: $H_{polar} = C_k \cdot S \cdot \cos(\pi H / 3), V_{polar} = C_k \cdot S \cdot \sin(\pi H / 3), I_{polar} = I_{max} = \max(R, G, B)$, where the adaptive intensity collapse factor is: $C_k = k \cdot \sin(\pi I_{max} / 2) + \epsilon$, with learnable density parameter $k$ initialized to 1.0 and constrained to $k \in [0.1, 5.0]$. This formulation eliminates red discontinuity through periodic parameterization, provides robustness through adaptive intensity collapse that prevents degenerate mappings in dark regions, and maintains color relationships under illumination changes. While \citep{yan2024you} \citep{yan2025hvi} frames \hyperref[ac]{HVI} as a representation transformation, we define an explicit color loss in \hyperref[ac]{HVI} space. The polarized color loss is computed as: $ L_{col} = \|H_{polar}^{pred} - H_{polar}^{gt}\|_1 + \|V_{polar}^{pred} - V_{polar}^{gt}\|_1 + \|I_{polar}^{pred} - I_{polar}^{gt}\|_1 $

\begin{figure*}[t]
    \centering
    \includegraphics[width=0.9\linewidth]{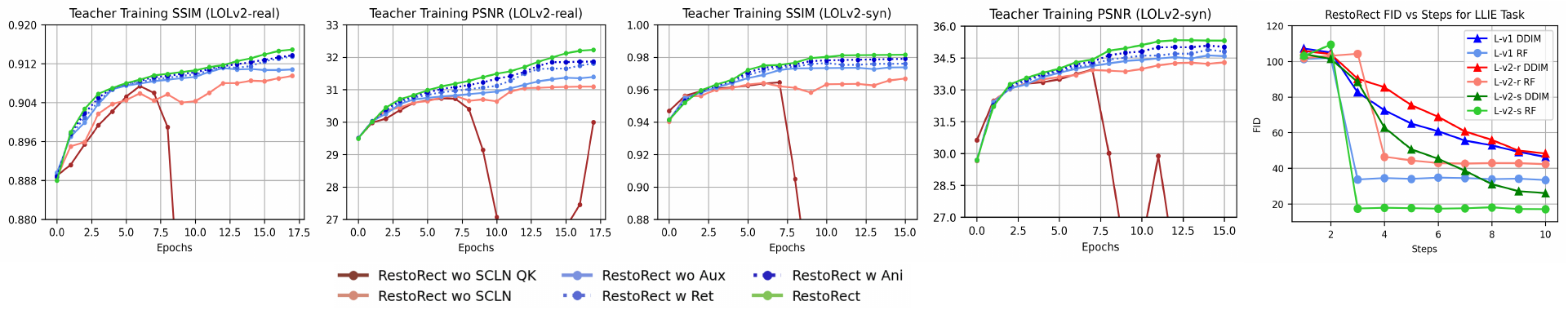}
    \caption{{\small(1-4)} Teacher model training with ablations of \hyperref[ac]{SCLN} \& \hyperref[ac]{QK} Norm (red) and auxiliary losses (blue). Model indicated by red line without \hyperref[ac]{SCLN} \& \hyperref[ac]{QK} Norm is identical to Reti-Diff \citep{he2023reti}. {\small(5)} FID vs Steps inference performance show Rectified Flow (RF) student model producing high quality images in fewer steps compared to Denoising Diffusion Implicit Model (\hyperref[ac]{DDIM}).}
    \label{fig:compare}
\end{figure*}

\begin{figure*}[h]
    \centering
    \includegraphics[width=0.9\linewidth]{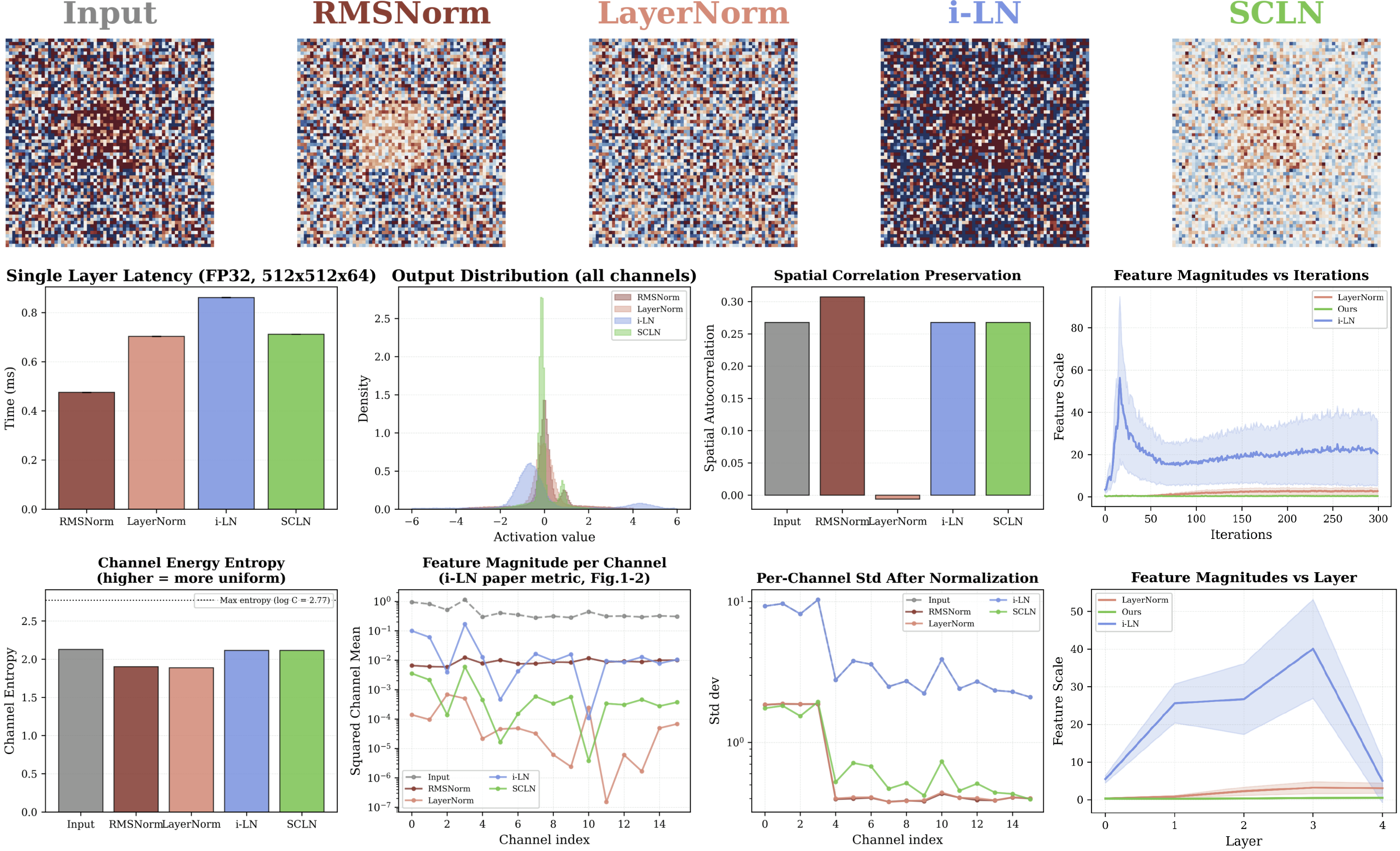}
    \caption{Comparison of normalization strategies for image restoration transformers. Evaluated SCLN against LayerNorm (per-token, the baseline used in Reti-Diff), RMSNorm (per-token, no mean subtraction), and i-LN (\cite{lee2025analyzing}, norm with global-std rescaling)}
    \label{fig:norm}
\end{figure*}

This novel combination ensures that the restored images maintain both structural accuracy and perceptual realism. The complete teacher training objective combines these losses: $
L_{teach} = L_{rec} + \lambda_{tex} L_{tex} + \lambda_{col} L_{col} $ with $\lambda_{tex} = 0.05$, $\lambda_{col} = 0.05$. Figure \ref{fig:compare} shows in green how auxiliary constraints allow training of a stronger teacher model with faster convergence. \citep{he2023reti} previously used reconstruction and style loss with perceptual \hyperref[ac]{VGG} features. To our knowledge, we are the first to implement anisotropic diffusion texture and illumination smoothness constraints with explicit \hyperref[ac]{HVI} color loss. Figure \ref{fig:compare} (1-4) Teacher model training ablations: dark red no SCLN, no QK norm (equivalent to Reti-Diff); light red SCLN only, no QK norm; dark blue dotted no anisotropic diffusion constraint; light blue dotted no polarized HVI color loss; green full RestoRect with all components. SCLN + QK norm together are required for stable training; removing either individually (light red) degrades convergence relative to the full system. Auxiliary constraints individually contribute, with their combination (green) yielding fastest convergence and best final quality.

\subsection{Student Network with Latent Rectified Flow}

Traditional knowledge distillation treats feature transfer as static matching between teacher and student representations, computing loss per sample and treating each teacher feature as a fixed pointwise regression target. This collapses the teacher's full feature distribution to a single point estimate per input, ignoring variance, cross-layer correlations, and multi-modal structure arising from multi-head attention over diverse degraded inputs. This approach suffers from several further limitations: it assumes compatible feature distributions between architectures, lacks flexibility in handling multi-modal feature distributions, and cannot adapt to varying restoration complexity. Our generative approach instead learns a velocity field over the distribution of teacher features. During training, the student samples noisy interpolants across the full ODE trajectory, exposing it to the distributional structure of teacher representations rather than converging to a single target per input. Trajectory consistency regularization explicitly enforces semantic coherence across integration steps, a property static matching cannot encode by construction. We reformulate knowledge distillation as a generative process using rectified flow, which models feature synthesis through straight-line paths in latent space.

Given teacher features $\mathbf{f}_{teach} \in \mathbb{R}^{d}$ and noise $\mathbf{z} \sim \mathcal{N}(0, I)$, rectified flow defines the interpolation path: $\mathbf{x}_t = (1-t) \mathbf{z} + t \mathbf{f}_{teach}, \quad t \in [0,1]$. The velocity field represents the direction of optimal transport: $\mathbf{v}(\mathbf{x}_t, t) = \frac{d\mathbf{x}_t}{dt} = \mathbf{f}_{teach} - \mathbf{z}$. We train separate velocity prediction networks $\epsilon_\theta^{rex}$ and $\epsilon_\theta^{img}$ for reflectance and image features using the velocity matching objective: $L_{vel} = \mathbb{E}_{t,\mathbf{z},\mathbf{f}_{teach}} \left[\|\epsilon_\theta(\mathbf{x}_t, t, \mathbf{c}) - \mathbf{v}(\mathbf{x}_t, t)\|_2^2\right]$, where $\mathbf{c}$ represents conditioning information from the input image. Each velocity predictor implements a Residual MLP architecture. During inference, we solve the \hyperref[ac]{ODE} using Euler's method with adaptive step sizing: $\mathbf{x}_{t+\Delta t} = \mathbf{x}_t + \Delta t \cdot \epsilon_\theta(\mathbf{x}_t, t, \mathbf{c})$. This requires only 1-4 integration steps compared to 10+ steps for \hyperref[ac]{DDIM} models, providing significant computational advantages. Standard knowledge distillation losses (KL divergence, L2 distance) assume that teacher and student features exist in compatible statistical distributions. This assumption fails for complex transformer architectures, and when fine-tuning on different datasets, leading to suboptimal knowledge transfer \citep{lin2022knowledge}.

\textbf{\hyperref[ac]{FLEX} (Feature Layer EXtraction) Loss} addresses feature distribution mismatch through cross-normalization for distribution alignment, percentile-based outlier detection for robust training, and dynamic resolution-aware weighting for multi-scale importance. Unlike \citep{berrada2025boosting} which is specialized for diffusion autoencoders, \hyperref[ac]{FLEX} provides a general-purpose distillation loss that transfers feature distributions across heterogeneous teacher-student architectures. The key method is cross-normalization using student statistics. For each layer $l$, \hyperref[ac]{FLEX} normalizes both teacher and student features using student statistics: $\mu_{\text{stud}}^l = \text{mean}(\mathbf{f}_{\text{stud}}^l), \quad \sigma_{\text{stud}}^l = \text{std}(\mathbf{f}_{\text{stud}}^l) + \epsilon, \quad \mathbf{f}_{\text{teach}}^{l,\text{norm}} = \frac{\mathbf{f}_{\text{teach}}^l - \mu_{\text{stud}}^l}{\sigma_{\text{stud}}^l}, \quad \mathbf{f}_{\text{stud}}^{l,\text{norm}} = \frac{\mathbf{f}_{\text{stud}}^l - \mu_{\text{stud}}^l}{\sigma_{\text{stud}}^l}$. This aligns both features to the student's distribution, enabling meaningful comparison across architecture capacity differences. \hyperref[ac]{FLEX} incorporates fast percentile-based outlier detection to handle extreme values that destabilize training. This masking strategy prioritizes training stability over complete spatial coverage, as extreme outliers generate destabilizing gradients that outweigh their informational value. The outlier mask identifies reliable spatial locations: $M_{\text{reliable}}^{l,c,h,w} = \mathbb{I}[|\mathbf{f}_{\text{stud}}^{l,c,\text{norm},h,w}| \leq \tau_p^{l,c}]$, where $\tau_p^{l,c}$ is the p-th percentile of normalized feature magnitudes for layer l, channel c, with p=95\% by default. \hyperref[ac]{FLEX} computes dynamic resolution-based weights: $w_l^{\text{res}} = \max\left(\left({H_{\text{base}} W_{\text{base}}}/{H_l W_l}\right)^{0.25}, 0.1\right)$, where $(H_{\text{base}}, W_{\text{base}}) = (64, 64)$ ensures appropriate weighting across resolutions. The complete \hyperref[ac]{FLEX} loss combines masked feature matching with dual weighting: $L_{\text{FLEX}} = \sum_{l} w_l^{\text{layer}} \cdot w_l^{\text{res}} \cdot \frac{\sum_{c,h,w} M_{\text{reliable}}^{l,c,h,w} \cdot \|\mathbf{f}_{\text{teach}}^{l,c,\text{norm},h,w} - \mathbf{f}_{\text{stud}}^{l,c,\text{norm},h,w}\|^2}{\sum_{c,h,w} M_{\text{reliable}}^{l,c,h,w} + \epsilon}$
where $w_l^{\text{layer}}$ represents predefined layer weights and the denominator normalizes by reliable elements. \hyperref[ac]{FLEX} includes SNR-aware application, activating only when $t/T < \tau_{\text{SNR}} = 0.4$, focusing distillation on cleaner intermediate states. Cross-normalization enables stable transfer between different architectures, outlier detection prevents training instability, dynamic weighting balances multi-scale contributions, and streaming processing optimizes memory usage. Standard KD methods lack these capabilities, assuming compatible distributions and uniform spatial weighting.

\textbf{Trajectory Consistency Regularization} is introduced to ensure smooth and semantically consistent rectified flow trajectories, which prevents erratic feature generation and maintains coherence throughout the \hyperref[ac]{ODE} integration process \citep{yang2024consistency}. We enforce smooth transitions between consecutive \hyperref[ac]{ODE} steps: $L_{trans} = \sum_{i=1}^{N-1} \|\mathbf{f}_{pred}^{i+1} - \mathbf{f}_{pred}^{i}\|_2^2$, where $\mathbf{f}_{pred}^{i}$ represents predicted features at the $i$-th integration step. We ensure final generated features align with teacher targets: $L_{target} = \|\mathbf{f}_{pred}^{final} - \mathbf{f}_{teach}\|_2^2$. We enforce consistency in semantic feature representations across the trajectory: $L_{cons} = \sum_{i=1}^{N} \text{cos\_dist}(\mathbf{f}_{pred}^{i}, \mathbf{f}_{teach})$. The complete trajectory consistency loss is: $L_{traj} = \alpha_{trans} L_{trans} + \alpha_{target} L_{target} + \alpha_{cons} L_{cons}$, with $\alpha_{trans} = 0.1$, $\alpha_{target} = 0.5$, and $\alpha_{cons} = 0.2$.
Our training protocol addresses the challenge of jointly learning velocity prediction and restoration quality through a principled two-phase approach. We first train rectified flow components while freezing the main restoration network:$ L_{phase1} = L_{vel}^{rex} + L_{vel}^{img} + \lambda_{KD} L_{KD} + \lambda_{traj} L_{traj}$. This phase establishes stable velocity prediction capabilities without interference from restoration objective gradients. We use separate optimizers for reflectance and image velocity predictors with learning rates $lr_{rex} = 2 \times 10^{-4}$ and $lr_{img} = 2 \times 10^{-4}$. The complete network is then trained using features generated by learned velocity predictors, where $\lambda_{FLEX} = 0.15$, $\lambda_{vel} = 0.05$: $ L_{phase2} = L_{rec} + \lambda_{FLEX} L_{FLEX} + \lambda_{vel} (L_{vel}^{rex} + L_{vel}^{img})$

\section{Experiments}

\textbf{Experimental Setup.} We implement our model in PyTorch and trained it on 8 NVIDIA H100 GPUs. Teacher pretraining is performed for 15-20 epochs depending on dataset convergence, while student phases I and II are each trained for 10 epochs. We use Adam optimizer with momentum terms (0.9, 0.999). For fair comparison with prior work \citep{he2023reti}, we adopt the same configuration of transformer blocks, attention heads, and channel dimensions: [3, 3, 3, 3], [1, 2, 4, 8], and [64, 128, 256, 512] from levels 1-4. During inference, we make 4 function evaluation calls for rectified flow generation, yielding faster generation and higher-quality outputs compared to state-of-the-art methods. Training follows the methodology of Reti-Diff and CIDNet across datasets and tasks.

\begin{table*}[t]
\setlength{\tabcolsep}{2pt}
\centering
\scriptsize
\caption{\hyperref[ac]{LLIE} task results. Best result shown in Green and second best shown in Blue.}
\label{tab:LLIE}
\vspace{-8pt}
\resizebox{0.80\textwidth}{!}{%
\begin{tabular}{l|cccc|cccc|cccc|cccc}
\toprule
\multirow{2}{*}{Methods}
& \multicolumn{4}{c|}{\hyperref[ac]{LOL}-v1} 
& \multicolumn{4}{c|}{\hyperref[ac]{LOL}-v2-real} 
& \multicolumn{4}{c|}{\hyperref[ac]{LOL}-v2-syn} 
& \multicolumn{4}{c}{\hyperref[ac]{SID}} \\ \cmidrule(lr){2-5} \cmidrule(lr){6-9} \cmidrule(lr){10-13} \cmidrule(lr){14-17}
& PSNR$\uparrow$ & SSIM$\uparrow$ & FID$\downarrow$ & BIQI$\downarrow$ 
& PSNR$\uparrow$ & SSIM$\uparrow$ & FID$\downarrow$ & BIQI$\downarrow$ 
& PSNR$\uparrow$ & SSIM$\uparrow$ & FID$\downarrow$ & BIQI$\downarrow$ 
& PSNR$\uparrow$ & SSIM$\uparrow$ & FID$\downarrow$ & BIQI$\downarrow$ \\ 
\midrule
MRQ \citep{liu2023low}        & 25.24 & 0.855 & 53.32 & 22.73 & 22.37 & 0.854 & 68.89 & 33.61 & 25.54 & 0.940 & 21.56 & 25.09 & 24.80 & 0.688 & 63.72 & 29.53 \\
IAGC \citep{wang2023low}      & 24.53 & 0.866 & 59.73 & 25.50 & 22.20 & 0.863 & 70.34 & 31.70 & 25.58 & 0.941 & 21.58 & 30.32 & 23.17 & 0.640 & 78.80 & 30.56 \\
DiffIR \citep{xia2023diffir}   & 23.15 & 0.828 & 70.13 & 26.38 & 21.15 & 0.816 & 72.33 & 29.15 & 24.76 & 0.921 & 21.36 & 27.74 & 23.17 & 0.640 & 78.80 & 30.56 \\
CUE \citep{zheng2023empowering} & 21.86 & 0.841 & 69.83 & 27.15 & 21.19 & 0.829 & 67.05 & 28.83 & 24.41 & 0.917 & 31.34 & 33.83 & 23.25 & 0.652 & 77.38 & 28.85 \\
Mamba \citep{guo2024mambair}   & 22.33 & 0.863 & 63.39 & 20.17 & 21.97 & 0.840 & 56.09 & 24.46 & 25.75 & \textbf{\textcolor{NavyBlue}{0.958}} & 17.95 & 20.37 & 21.14 & 0.656 & 154.76 & 32.72 \\
RetiDiff \citep{he2023reti}    & \textbf{\textcolor{NavyBlue}{25.35}} & 0.866 & 49.14 & 17.75 & 22.97 & 0.858 & \textbf{\textcolor{NavyBlue}{43.18}} & 23.66 & \textbf{\textcolor{NavyBlue}{27.53}} & 0.951 & \textbf{\textcolor{Green}{13.82}} & \textbf{\textcolor{NavyBlue}{15.77}} & \textbf{\textcolor{NavyBlue}{25.53}} & \textbf{\textcolor{NavyBlue}{0.692}} & \textbf{\textcolor{Green}{51.66}} & \textbf{\textcolor{NavyBlue}{25.58}} \\
CIDNet \citep{yan2024you} & 23.50 & \textbf{\textcolor{NavyBlue}{0.900}} & \textbf{\textcolor{NavyBlue}{46.69}} & \textbf{\textcolor{NavyBlue}{14.77}} & \textbf{\textcolor{Green}{24.11}} & \textbf{\textcolor{NavyBlue}{0.871}} & 48.04 & \textbf{\textcolor{NavyBlue}{18.45}} & 25.71 & 0.942 & 18.60 & 15.87 & 22.90 & 0.676 & 55.29 & 29.12 \\
\midrule
RestoRect (teacher only) & 22.18 & 0.862 & 63.77 & 26.50 & 20.11 & 0.833 & 65.84 & 29.21 & 23.15 & 0.911 & 28.72 & 28.13 & 22.60 & 0.717 & 68.42 & 27.13 \\
RestoRect & \textbf{\textcolor{Green}{27.84}} & \textbf{\textcolor{Green}{0.945}} & \textbf{\textcolor{Green}{38.67}} & \textbf{\textcolor{Green}{8.35}} & \textbf{\textcolor{NavyBlue}{22.97}} & \textbf{\textcolor{Green}{0.911}} & \textbf{\textcolor{Green}{42.80}} & \textbf{\textcolor{Green}{10.47}} & \textbf{\textcolor{Green}{27.69}} & \textbf{\textcolor{Green}{0.968}} & \textbf{\textcolor{NavyBlue}{16.75}} & \textbf{\textcolor{Green}{11.67}} & \textbf{\textcolor{Green}{26.19}} & \textbf{\textcolor{Green}{0.923}} & \textbf{\textcolor{NavyBlue}{54.23}} & \textbf{\textcolor{Green}{19.57}} \\
\bottomrule
\end{tabular}%
}
\end{table*}

\begin{figure*}[h]
    \centering
    \includegraphics[width=0.92\linewidth]{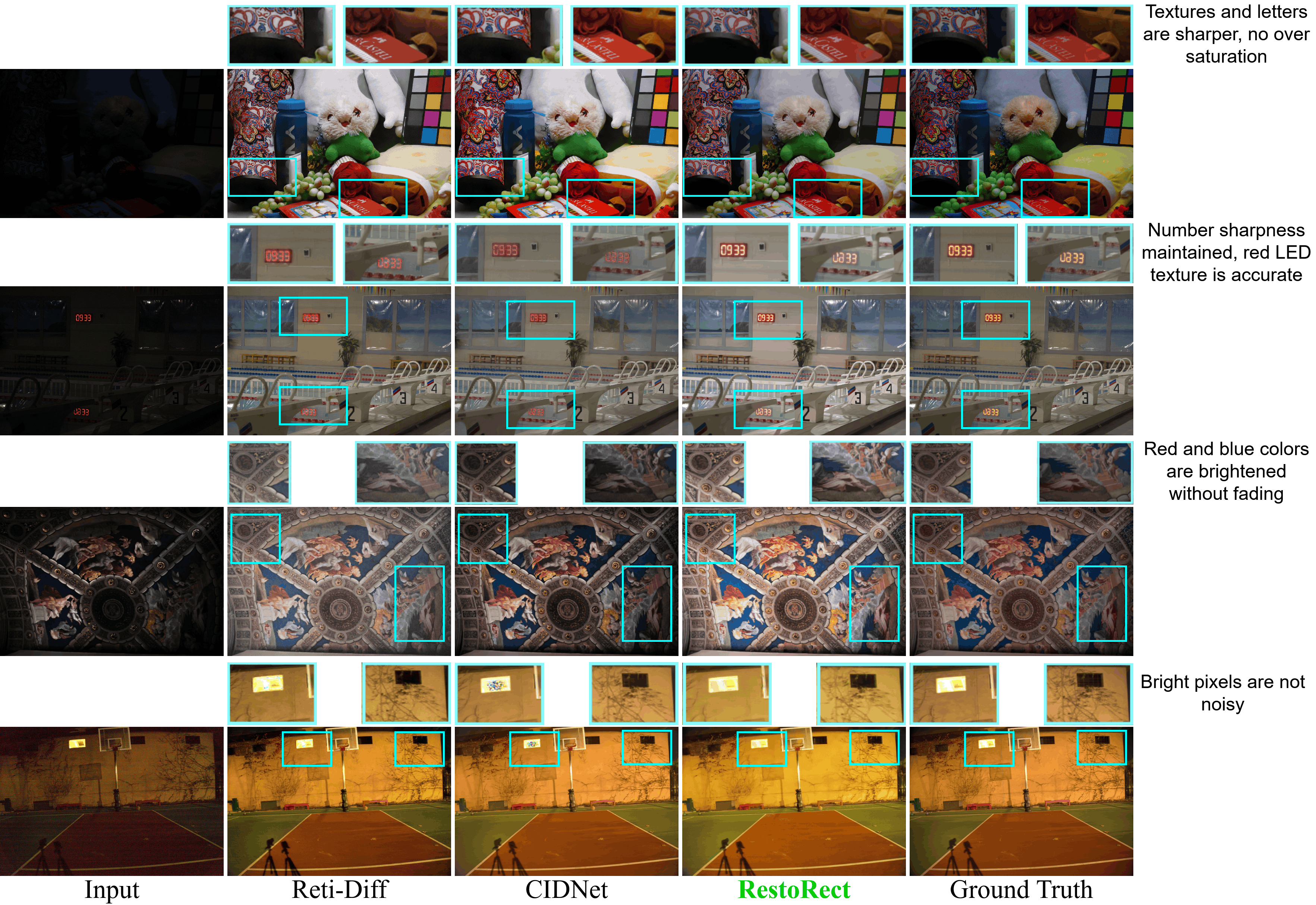}
    \caption{\hyperref[ac]{LLIE} task visual results (Top to Bottom: \hyperref[ac]{LOL}-v1, v2-real, v2-syn, \hyperref[ac]{SID})}
    \label{fig:lolfig}
\end{figure*}

\begin{table*}[t]
\setlength{\tabcolsep}{2pt}
\centering
\begin{minipage}[t]{0.438\textwidth}
\centering
\scriptsize
\caption{\hyperref[ac]{UIE} task results}
\vspace{-8pt}
\label{tab:UIE}
\resizebox{\textwidth}{!}{%
\begin{tabular}{l|ccc|ccc}
\toprule
\multirow{2}{*}{Methods}
& \multicolumn{3}{c|}{UIEB} & \multicolumn{3}{c}{\hyperref[ac]{LSUI}} \\
\cmidrule(lr){2-4} \cmidrule(lr){5-7}
& PSNR$\uparrow$ & SSIM$\uparrow$ & UIQM$\uparrow$
& PSNR$\uparrow$ & SSIM$\uparrow$ & UIQM$\uparrow$ \\
\midrule
SUnwet \citep{naik2021shallow} & 18.28 & 0.855 & 2.942 & 20.89 & 0.875 & 2.746 \\
PUIE \citep{fu2022uncertainty} & 21.38 & 0.882 & 3.021 & 23.70 & 0.902 & 2.974 \\
UShape \citep{peng2023u}       & 22.91 & 0.905 & 2.896 & 24.16 & 0.917 & 3.022 \\
PUGAN \citep{cong2023pugan}    & 23.05 & 0.897 & 2.902 & 25.06 & 0.916 & 3.106 \\
ADP \citep{zhou2023underwater} & 22.90 & 0.892 & 3.005 & 24.28 & 0.913 & 3.075 \\
NU2Net \citep{guo2023underwater} & 22.38 & 0.903 & 2.936 & 25.07 & 0.908 & 3.112 \\
AST \citep{zhou2024adapt}      & 22.19 & 0.908 & 2.981 & 27.46 & 0.916 & 3.107 \\
Mamba \citep{guo2024mambair}   & 22.60 & \textbf{\textcolor{NavyBlue}{0.939}} & 2.991 & 27.68 & 0.916 & 3.118 \\
RetiDiff \citep{he2023reti}    & \textbf{\textcolor{NavyBlue}{24.12}} & 0.910 & \textbf{\textcolor{NavyBlue}{3.088}} & \textbf{\textcolor{NavyBlue}{28.10}} & \textbf{\textcolor{NavyBlue}{0.929}} & \textbf{\textcolor{NavyBlue}{3.208}} \\
\midrule
RestoRect & \textbf{\textcolor{Green}{25.88}} & \textbf{\textcolor{Green}{0.950}} & \textbf{\textcolor{Green}{3.121}} & \textbf{\textcolor{Green}{28.10}} & \textbf{\textcolor{Green}{0.937}} & \textbf{\textcolor{Green}{3.229}} \\
\bottomrule
\end{tabular}
}
\end{minipage}%
\hfill
\begin{minipage}[t]{0.2975\textwidth}
\centering
\scriptsize
\caption{\hyperref[ac]{BAID} task results}
\vspace{-8pt}
\label{tab:BAID}
\resizebox{\textwidth}{!}{%
\begin{tabular}{|l|cccc}
\toprule
\multirow{2}{*}{Methods} & \multicolumn{3}{c}{BAID} \\
\cmidrule(lr){2-4}
& PSNR$\uparrow$ & SSIM$\uparrow$ & FID$\downarrow$ \\
\midrule
EnGAN \citep{jiang2021enlightengan} & 17.96 & 0.819 & 43.55 \\
URetinex \citep{wu2022uretinex} & 19.08 & 0.845 & 42.26 \\
CLIPLIT \citep{liang2023iterative} & 21.13 & 0.853 & 37.30 \\
DiffRet \citep{yi2023diff} & 22.07 & 0.861 & 38.07 \\
DiffIR \citep{xia2023diffir} & 21.10 & 0.835 & 40.35 \\
AST \citep{zhou2024adapt} & 22.61 & 0.851 & 32.47 \\
Mamba \citep{guo2024mambair} & 23.07 & 0.874 & 29.13 \\
RAVE \citep{gaintseva2024rave} & 21.26 & 0.872 & 64.89 \\
RetiDiff \citep{he2023reti} & \textbf{\textcolor{NavyBlue}{23.19}} & \textbf{\textcolor{NavyBlue}{0.876}} & \textbf{\textcolor{NavyBlue}{27.47}} \\
\midrule
RestoRect & \textbf{\textcolor{Green}{27.67}} & \textbf{\textcolor{Green}{0.965}} & \textbf{\textcolor{Green}{15.82}} \\
\bottomrule
\end{tabular}
}
\end{minipage}%
\hfill
\begin{minipage}[t]{0.2635\textwidth}
\centering
\scriptsize
\caption{\hyperref[ac]{FIE} task results}
\vspace{-8pt}
\label{tab:FIE}
\resizebox{\textwidth}{!}{%
\begin{tabular}{|l|ccc}
\toprule
\multirow{2}{*}{Methods} & \multicolumn{2}{c}{Fundus} \\
\cmidrule(lr){2-3}
& BIQI$\downarrow$ & CLIPQ$\uparrow$ \\
\midrule
SNRNet \citep{xu2022snr} & \textbf{\textcolor{NavyBlue}{6.144}} & \textbf{\textcolor{NavyBlue}{0.557}} \\
URetinex \citep{wu2022uretinex} & 12.158 & \textbf{\textcolor{Green}{0.561}} \\
SCI \citep{ma2022toward} & 23.527 & 0.552 \\
MIRNet \citep{zamir2022learning} & 14.925 & 0.527 \\
FourLLE \citep{wang2023fourllie} & 7.741 & 0.508 \\
CUE \citep{zheng2023empowering}  & 11.721 & 0.448 \\
NeRCO \citep{yang2023implicit}   & 17.256 & 0.451 \\
RetiDiff \citep{he2023reti}  & 10.788 & 0.525 \\
CIDNet \citep{yan2024you} & 10.663 & 0.529 \\
\midrule
RestoRect & \textbf{\textcolor{Green}{6.033}} & 0.503 \\
\bottomrule
\end{tabular}
}
\end{minipage}
\end{table*}

\begin{figure*}[h]
    \centering
    \includegraphics[width=1\linewidth]{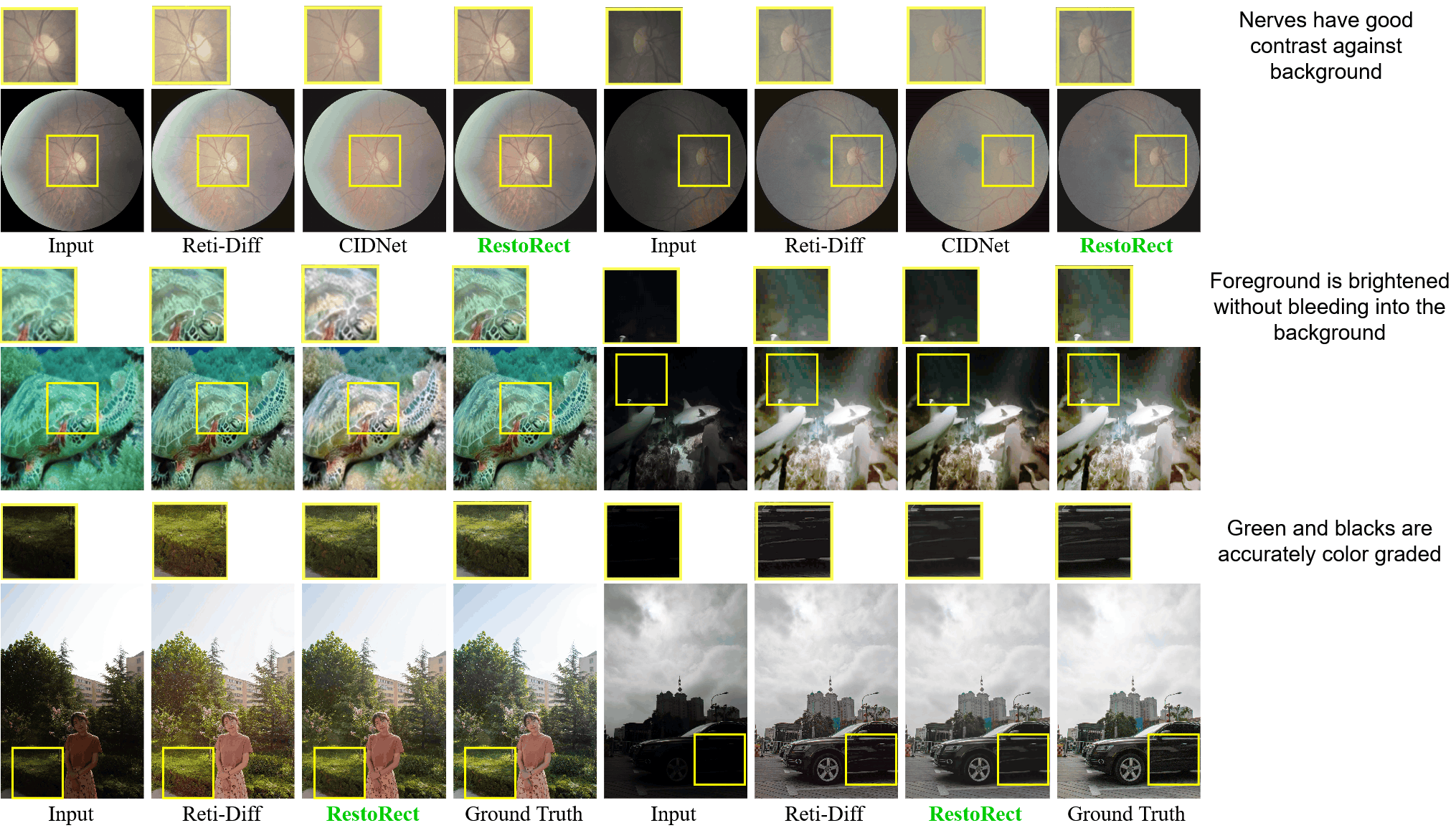}
    \caption{\hyperref[ac]{FIE} (Top), \hyperref[ac]{UIEB} (Middle Left), \hyperref[ac]{LSUI} (Middle Right), \hyperref[ac]{BAID} (Bottom) task visual results}
    \label{fig:uiefig}
\end{figure*}

\begin{table*}[t]
\setlength{\tabcolsep}{2pt}
\centering
\begin{minipage}[t]{0.335\textwidth}
\centering
\scriptsize
\caption{Unpaired task results}
\vspace{-8pt}
\label{tab:unpaired}
\resizebox{\textwidth}{!}{%
\begin{tabular}{l|ccccc|}
\toprule
\multirow{2}{*}{Methods} & \hyperref[ac]{DCIM} & \hyperref[ac]{LIME} & \hyperref[ac]{MEF} & \hyperref[ac]{NPE} & \hyperref[ac]{VV} \\
\cmidrule(lr){2-6}
& \multicolumn{5}{c|}{BRISQUE$\downarrow$} \\
\midrule
KinD \citep{zhang2019kindling} & 48.72 & 39.91 & 49.94 & 36.85 & 50.56 \\
ZeroDCE \citep{guo2020zero} & 27.56 & 20.44 & 17.32 & 24.72 & 34.66 \\
RUAS \citep{liu2021retinex} & 38.75 & 27.59 & 23.68 & 47.85 & 38.37 \\
LLFlow \citep{wang2022low} & 26.36 & 27.06 & 30.27 & 28.86 & 31.67 \\
SNRAware \citep{xu2022snr} & 37.35 & 39.22 & 31.28 & 26.65 & 78.72 \\
PairLIE \citep{fu2023learning} & 33.31 & 25.23 & 27.53 & 28.27 & 39.13 \\
CIDNet \citep{yan2024you} & \textbf{\textcolor{NavyBlue}{21.47}} & \textbf{\textcolor{NavyBlue}{16.25}} & \textbf{\textcolor{Green}{13.77}} & \textbf{\textcolor{Green}{18.92}} & \textbf{\textcolor{NavyBlue}{30.63}} \\
\midrule
RestoRect    & \textbf{\textcolor{Green}{16.56}} & \textbf{\textcolor{Green}{16.12}} & \textbf{\textcolor{NavyBlue}{14.69}} & \textbf{\textcolor{NavyBlue}{23.91}} & \textbf{\textcolor{Green}{24.42}} \\
\bottomrule
\end{tabular}
}
\end{minipage}%
\hfill
\begin{minipage}[t]{0.422\textwidth}
\centering
\scriptsize
\caption{\hyperref[ac]{SICE} task results}
\vspace{-8pt}
\label{tab:SICE}
\resizebox{\textwidth}{!}{%
\begin{tabular}{l|ccc|ccc|}
\toprule
\multirow{2}{*}{Methods} & \multicolumn{3}{c|}{\hyperref[ac]{SICE}-Mix} & \multicolumn{3}{c|}{\hyperref[ac]{SICE}-Grad} \\
\cmidrule(lr){2-4} \cmidrule(lr){5-7}
& PSNR$\uparrow$ & SSIM$\uparrow$ & LPIPS$\downarrow$ & PSNR$\uparrow$ & SSIM$\uparrow$ & LPIPS$\downarrow$ \\
\midrule
RetiNet \citep{wei2018deep} & 12.397 & 0.606 & 0.407 & 12.450 & 0.619 & 0.364 \\
ZeroDCE \citep{guo2020zero} & 12.428 & 0.633 & \textbf{\textcolor{NavyBlue}{0.382}} & 12.475 & 0.644 & \textbf{\textcolor{NavyBlue}{0.334}} \\
URetinex \citep{wu2022uretinex} & 10.903 & 0.600 & 0.402 & 10.894 & 0.610 & 0.356 \\
RUAS \citep{liu2021retinex} & 8.684 & 0.493 & 0.525 & 8.628 & 0.494 & 0.499 \\
LLFlow \citep{wang2022low} & 12.737 & 0.617 & 0.388 & 12.737 & 0.617 & 0.388 \\
LEDNet \citep{zhou2022lednet} & 12.668 & 0.579 & 0.412 & 12.551 & 0.576 & 0.383 \\
CIDNet \citep{yan2024you} & \textbf{\textcolor{NavyBlue}{13.425}} & \textbf{\textcolor{NavyBlue}{0.636}} & \textbf{\textcolor{Green}{0.362}} & \textbf{\textcolor{NavyBlue}{13.446}} & \textbf{\textcolor{NavyBlue}{0.648}} & \textbf{\textcolor{Green}{0.318}} \\
\midrule
RestoRect & \textbf{\textcolor{Green}{15.041}} & \textbf{\textcolor{Green}{0.667}} & 0.393 & \textbf{\textcolor{Green}{15.447}} & \textbf{\textcolor{Green}{0.715}} & 0.354 \\
\bottomrule
\end{tabular}
}
\end{minipage}
\hfill
\begin{minipage}[t]{0.236\textwidth}
\centering
\scriptsize
\caption{Generalizability}
\vspace{-8pt}
\label{tab:ablation}
\resizebox{\textwidth}{!}{%
\begin{tabular}{lc|cccc}
\toprule
Test & Train & PSNR$\uparrow$ & SSIM$\uparrow$ & FID$\downarrow$ & BIQI$\downarrow$ \\
\midrule
\multirow{3}{*}{v1} 
& {\tiny-FLEX} & 24.27 & 0.891 & 44.75 & 9.02 \\
& v2-s & 18.32 & 0.827 & 99.36 & 18.74 \\
& v2-r & 17.57 & 0.827 & 111.66 & 21.68 \\
\midrule
\multirow{3}{*}{v2-r} 
& {\tiny-FLEX} & 23.16 & 0.880 & 41.55 & 10.52 \\
& v1 & 22.27 & 0.874 & 48.92 & 18.57 \\
& v2-s & 21.15 & 0.837 & 106.29 & 22.92 \\
\midrule
\multirow{3}{*}{v2-s} 
& {\tiny-FLEX} & 27.89 & 0.942 & 17.93 & 11.95 \\
& v1 & 19.96 & 0.876 & 69.37 & 16.39 \\
& v2-r & 17.18 & 0.768 & 117.84 & 25.26 \\
\bottomrule
\end{tabular}
}
\end{minipage}
\end{table*}

\begin{figure*}[h]
    \centering
    \includegraphics[width=0.90\linewidth]{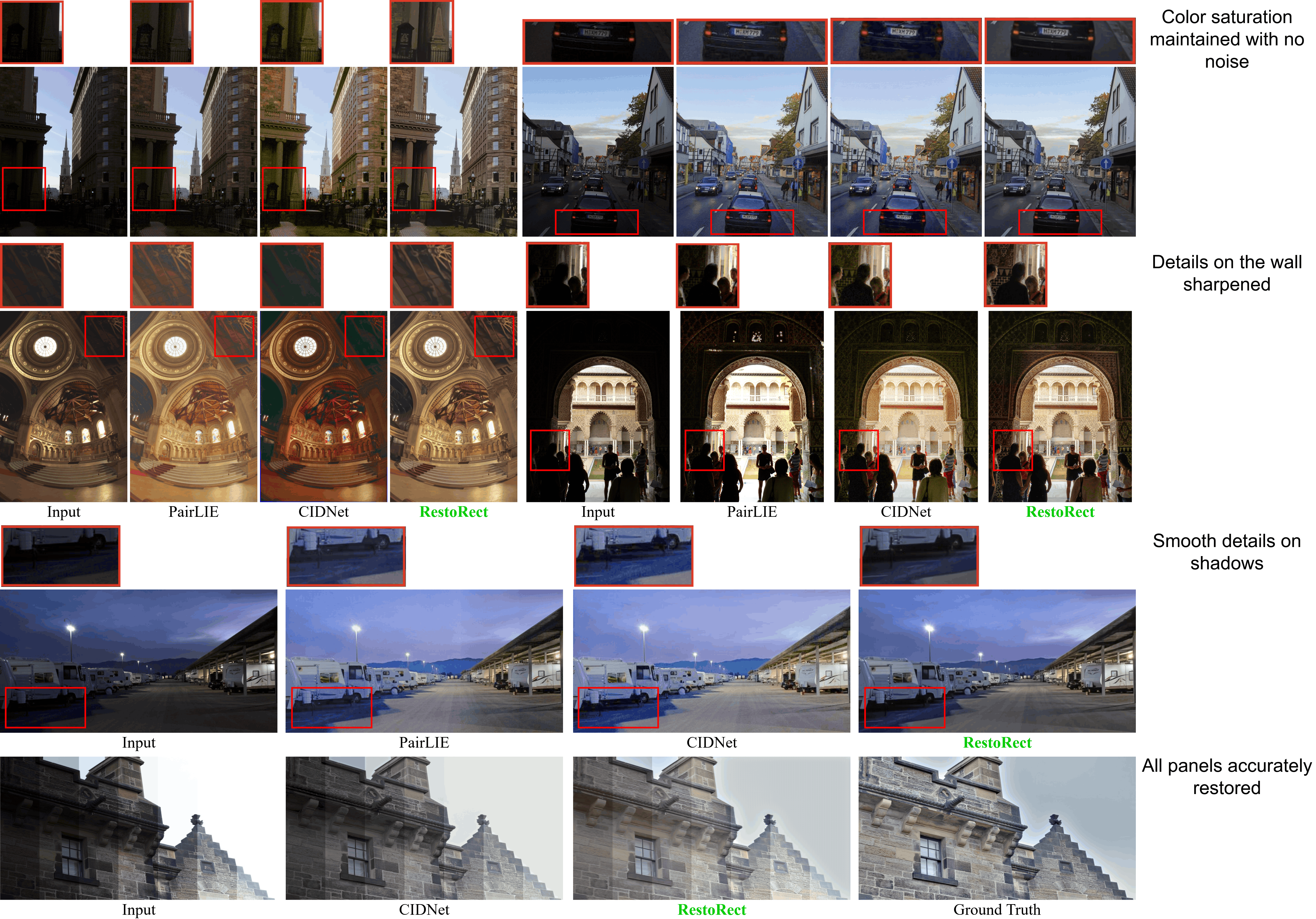}
    \caption{\hyperref[ac]{DCIM} (Row 1 Left), \hyperref[ac]{LIME} (Row 1 Right), \hyperref[ac]{MEF} (Row 2 Left), \hyperref[ac]{NPE} (Row 2 Right), \hyperref[ac]{VV} (Row 3), \hyperref[ac]{SICE}-Grad (Row 4) results}
    \label{fig:unpairedfig}
\end{figure*}

\begin{figure*}[t]
    \centering
    \includegraphics[width=0.99\linewidth]{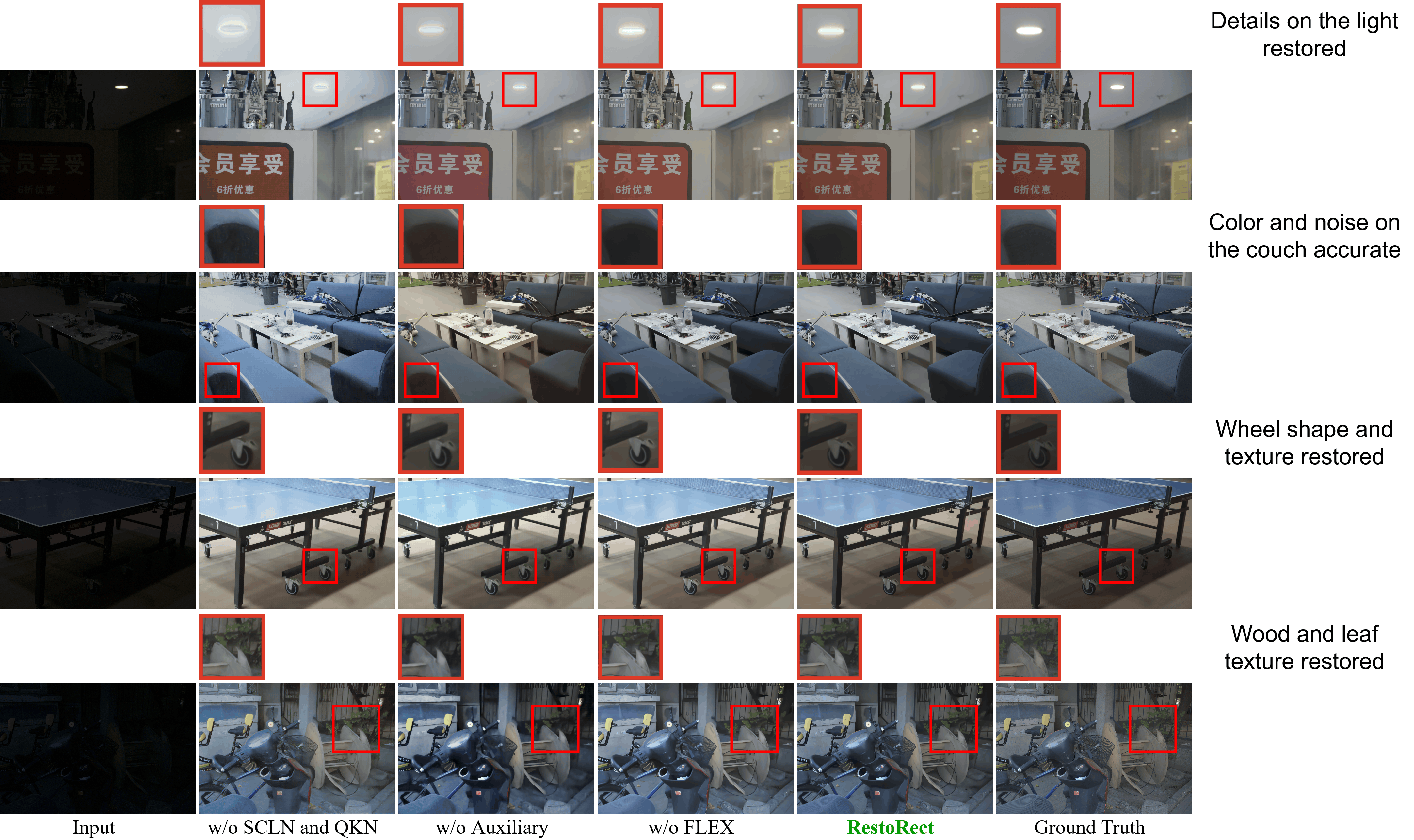}
    \caption{\hyperref[ac]{LLIE} task visual results with ablation of \hyperref[ac]{SCLN} and \hyperref[ac]{QK} Norm, Retinex Priors, \hyperref[ac]{FLEX} loss, compared to full RestoRect architecture and Ground Truth.}
    \label{fig:ablation}
\end{figure*}

\begin{table*}[t]
\centering
\begin{minipage}[t]{0.67\textwidth}
\setlength{\tabcolsep}{2pt}
\caption{Downstream application on low-light image detection task on ExDark dataset.}
\label{downstream-task}
\vspace{-8pt}
\resizebox{\textwidth}{!}{
\begin{tabular}{l|cccccccccccc|c}
\toprule
Methods & Bicycle & Boat & Bottle & Bus & Car & Cat & Chair & Cup & Dog & Motor & People & Table & Mean \\
\midrule
Restormer   & 77.0 & 71.0 & 68.8 & 91.6 & 77.1 & 62.5 & 57.3 & 68.0 & 69.6 & 69.2 & 74.6 & 49.7 & 69.7 \\
SCI         & 73.4 & 68.0 & 69.5 & 86.2 & 74.5 & 63.1 & 59.5 & 61.0 & 67.3 & 63.9 & 73.2 & 47.3 & 67.2 \\
SNR-Net     & 78.3 & 74.2 & 74.5 & 89.6 & 82.7 & 66.8 & 66.3 & 62.5 & 74.7 & 63.1 & 73.3 & 57.2 & 71.9 \\
Retformer   & 78.1 & 74.5 & 74.2 & 91.2 & 82.2 & 65.0 & 63.3 & 67.0 & 75.4 & 68.6 & 75.3 & 55.6 & 72.5 \\
RetiDiff    & \textbf{\textcolor{NavyBlue}{82.0}} & \textbf{\textcolor{NavyBlue}{77.9}} & \textbf{\textcolor{NavyBlue}{76.4}} & \textbf{\textcolor{Green}{92.2}} & \textbf{\textcolor{NavyBlue}{83.3}} & \textbf{\textcolor{NavyBlue}{69.6}} & \textbf{\textcolor{NavyBlue}{67.4}} & \textbf{\textcolor{Green}{74.4}} & \textbf{\textcolor{NavyBlue}{75.5}} & \textbf{\textcolor{NavyBlue}{74.3}} & \textbf{\textcolor{Green}{78.3}} & \textbf{\textcolor{Green}{57.9}} & \textbf{\textcolor{NavyBlue}{75.8}} \\
\midrule
RestoRect   & \textbf{\textcolor{Green}{85.8}} & \textbf{\textcolor{Green}{79.1}} & \textbf{\textcolor{Green}{79.5}} & \textbf{\textcolor{NavyBlue}{91.5}} & \textbf{\textcolor{Green}{83.5}} & \textbf{\textcolor{Green}{70.2}} & \textbf{\textcolor{Green}{68.8}} & \textbf{\textcolor{NavyBlue}{74.1}} & \textbf{\textcolor{Green}{78.2}} & \textbf{\textcolor{Green}{77.9}} & \textbf{\textcolor{NavyBlue}{78.2}} & \textbf{\textcolor{NavyBlue}{57.4}} & \textbf{\textcolor{Green}{77.1}} \\
\bottomrule
\end{tabular}}
\end{minipage}
\hfill
\begin{minipage}[t]{0.32\textwidth}
\setlength{\tabcolsep}{2pt}
\caption{Model complexity comparison}
\label{tab:complexity}
\vspace{-8pt}
\centering
\resizebox{\textwidth}{!}{
\begin{tabular}{l|c|cc}
\toprule
Methods & Source & Params M & GFLOPs \\
\midrule
Diff-Reti & ICCV'23 & 56.88 & 198.16 \\
DiffIR & ICCV'23 & 27.80 & 35.32 \\
GSAD & NIPS'23 & 17.17 & 670.33 \\
Reti-Diff & ICLR'25 & 26.11 & 87.63 \\
\midrule
RestoRect & teacher & 27.98 & 53.20 \\
RestoRect & student & 25.87 & 49.50 \\
\bottomrule
\end{tabular}
}
\end{minipage}
\end{table*}

\textbf{Quantitative Evaluation.} For the low-light image enhancement (LLIE) task, we conduct experiments on \hyperref[ac]{LOL}-v1 \citep{wei2018deep}, \hyperref[ac]{LOL}-v2-real, \hyperref[ac]{LOL}-v2-syn \citep{yang2021sparse}, and \hyperref[ac]{SID} \citep{chen2019seeing}. RestoRect achieves state-of-the-art performance across all datasets shown in Table \ref{tab:LLIE}, with improvements on almost every metric over the second-best methods (RetiDiff and CIDNet). The visual results shown in Figure \ref{fig:lolfig} highlight clear improvements in fine grained details shown in cyan boxes (please zoom in for clarity).

For the underwater image enhancement (UIE) task, we evaluate on \hyperref[ac]{UIEB} \citep{li2019underwater} and \hyperref[ac]{LSUI} \citep{peng2023u}, where RestoRect outperforms RetiDiff by 1.76dB PSNR on \hyperref[ac]{UIEB} and matches its performance on \hyperref[ac]{LSUI} while achieving superior SSIM scores shown in Table \ref{tab:UIE}. For the backlit image enhancement (BAID) task, experiments are performed on \citep{lv2022backlitnet} where RestoRect demonstrates substantial improvements with 4.48dB PSNR gain over RetiDiff and 11.65 FID reduction shown in Table \ref{tab:BAID}. Additionally, we test on real-world fundus image enhancement (FIE) \citep{deep_reitna_enhance} images using the \hyperref[ac]{LOL}-v2-syn pretrained model, where RestoRect achieves the lowest BIQI score of 6.033, outperforming SNRNet shown in Table \ref{tab:FIE}. The visual results shown in Figure \ref{fig:uiefig} highlight our performance with details shown in yellow boxes (please zoom in for clarity). Reti-Diff baseline images for \hyperref[ac]{UIEB} and \hyperref[ac]{LSUI} in middle row very closely match the ground truth while the scores are marginally worse than ours. This makes us believe that the publicly available checkpoints provided on Reti-Diff's github might be overfitted to the validation set, unlike our model which has not seen the validation set.

For real-world image restoration, we test on five unpaired datasets: \hyperref[ac]{DCIM} \citep{lee2013contrast}, \hyperref[ac]{LIME} \citep{guo2016lime}, \hyperref[ac]{MEF} \citep{wang2013naturalness}, \hyperref[ac]{NPE} \citep{ma2015perceptual}, and \hyperref[ac]{VV} \citep{he2025diffusion}. Using the \hyperref[ac]{LOL}-v2-syn pretrained model for inference, where RestoRect consistently outperforms CIDNet across most datasets, achieving the best scores on \hyperref[ac]{DCIM} (16.56), \hyperref[ac]{LIME} (16.12), and \hyperref[ac]{VV} (24.42) as shown in Table \ref{tab:unpaired}. We further evaluate on single image contrast enhancement (\hyperref[ac]{SICE}) \citep{cai2018learning}, which contains underexposed and overexposed images, training on the resized \hyperref[ac]{SICE} training set and test on the datasets \hyperref[ac]{SICE}-Mix and \hyperref[ac]{SICE}-Grad \citep{zheng2022low}. RestoRect achieves superior PSNR and SSIM performance over CIDNet by 1.6dB and 0.031 on \hyperref[ac]{SICE}-Mix, and 2.0dB and 0.077 on \hyperref[ac]{SICE}-Grad, as shown in Table \ref{tab:SICE}.

\begin{figure*}[t]
\centering
\includegraphics[width=1\linewidth]{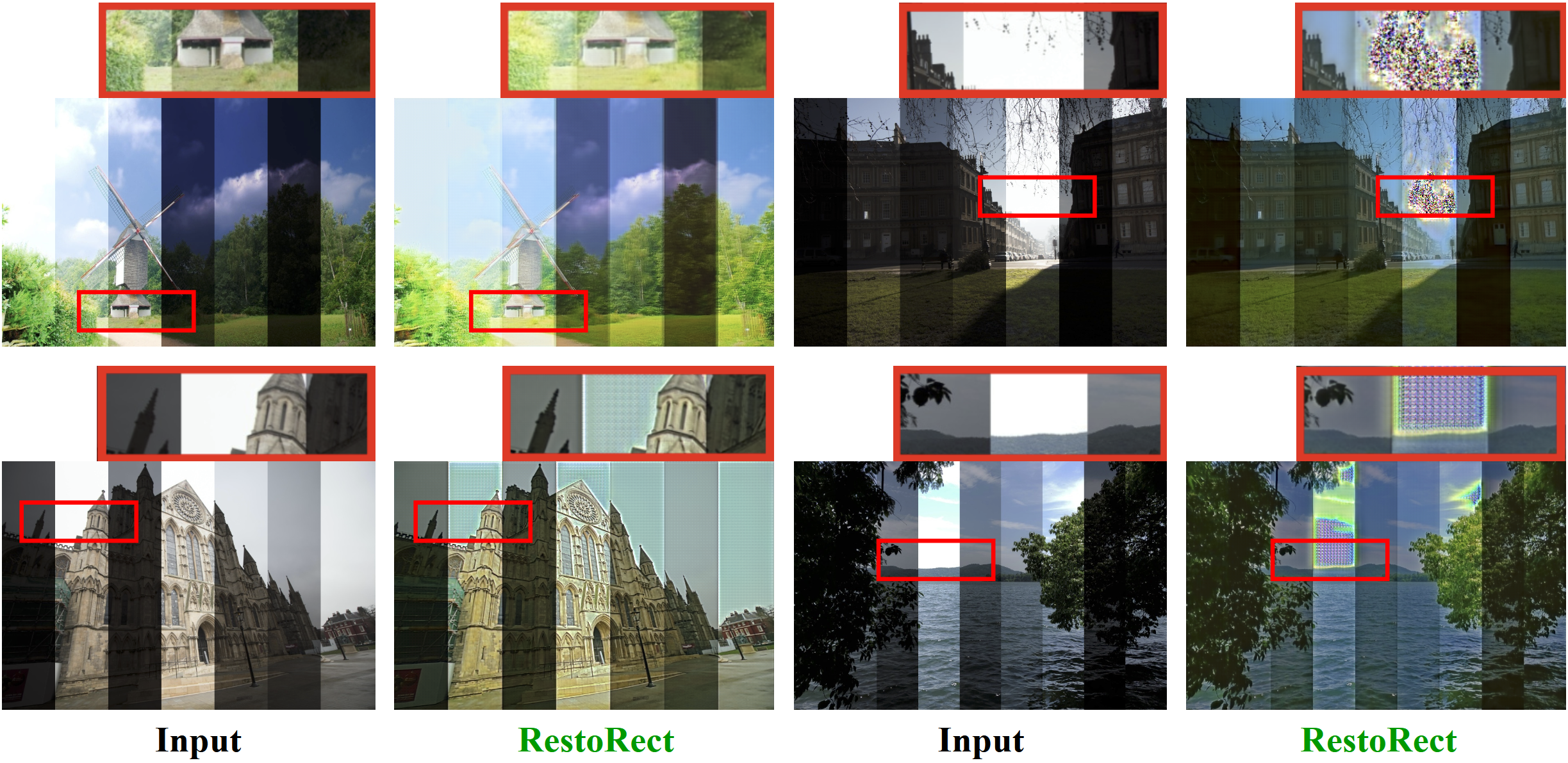}
\caption{Failure cases on \hyperref[ac]{SICE}-Mix dataset.}
\label{fig:failure}
\end{figure*}

\begin{table}[t]
\centering
\caption{Student ablation study (L-v1).}
\label{tab:student_ablation}
\vspace{-8pt}
\resizebox{\columnwidth}{!}{
\setlength{\tabcolsep}{2pt}
\begin{tabular}{lcccc}
\toprule
Configuration & PSNR$\uparrow$ & SSIM$\uparrow$ & FID$\downarrow$ & BIQI$\downarrow$ \\
\midrule
RetiDiff (Traditional KD) & 25.35 & 0.866 & 49.14 & 17.75\\
\midrule
1-Vel-Net RF(TC) + no FLEX & 21.10 & 0.762 & 52.19 & 20.95\\
1-Vel-Net RF(TC) + FLEX & 21.52 & 0.791 & 51.24 & 20.88\\
\midrule
RF(no TC) + no FLEX & 24.10 & 0.855 & 49.28 & 17.95\\
RF(TC) + no FLEX & 24.27 & 0.891 & 44.75 & 9.02\\
RF(TC) + FLEX(CN) & 25.64 & 0.913 & 42.18 & 8.73\\
RF(TC) + FLEX(CN+PM) & 27.82 & \textbf{0.947} & 39.25 & 8.42\\
RF(TC) + FLEX(CN+PM+RW) & \textbf{27.84} & 0.945 & \textbf{38.67} & \textbf{8.35}\\
\midrule
Percentile Threshold $p = 85\%$ & 26.21 & 0.898 & 39.84 & 8.58\\
Percentile Threshold $p = 90\%$ & 27.58 & \textbf{0.946} & 39.12 & 8.43\\
Percentile Threshold $p = 95\%$ & \textbf{27.84} & 0.945 & \textbf{38.67} & \textbf{8.35}\\
Percentile Threshold $p = 99\%$ & 27.67 & 0.943 & 38.91 & 8.41\\
\midrule
SNR Threshold $\tau_{SNR} = 0.2$ & 27.15 & 0.937 & 40.12 & 8.64\\
SNR Threshold $\tau_{SNR} = 0.4$ & 27.84 & \textbf{0.945} & \textbf{38.67} & \textbf{8.35}\\
SNR Threshold $\tau_{SNR} = 0.6$ & \textbf{27.89} & 0.941 & 39.28 & 8.47\\
SNR Threshold $\tau_{SNR} = 0.8$ & 26.92 & 0.934 & 40.53 & 8.71\\
\bottomrule
\end{tabular}
}
\end{table}

\begin{figure}[t]
    \centering
    \includegraphics[trim={0 0 0 0},clip,width=1\linewidth]{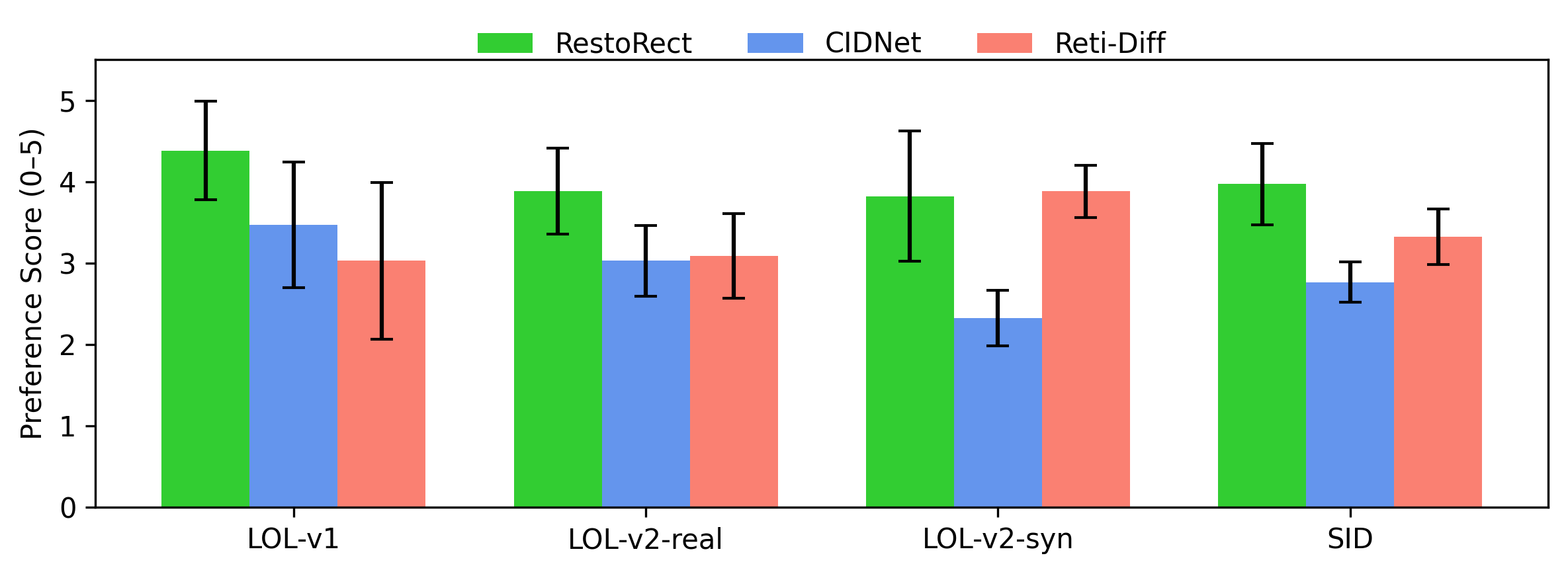}
    \caption{Qualitative human evaluation user study with 17 participants on \hyperref[ac]{LLIE} datasets.}
    \label{fig:eval}
\end{figure}

\textbf{Qualitative Evaluation.} We conduct a user study to evaluate low-light image enhancement. 17 participants (university computer vision researchers and students) are shown 20 low-light images alongside enhanced outputs from RestoRect, Reti-Diff, and CIDNet (RAVE included for BAID dataset). In a blind comparison, subjects rate each result on a 5-point scale for closeness to the ground truth. Figure \ref{fig:eval} presents the preference distributions, showing that RestoRect consistently achieves the highest preference across all five datasets, highlighting its ability to generate visually appealing results perceived as closest to the ground truth. Figure \ref{fig:eval} and Table \ref{tab:complexity} shows comparison of RestoRect's student model parameter size (M) and GFLOPs against other transformer architecture baselines demonstrating efficiency. The teacher network is used only during training to provide feature distribution targets and is never deployed at inference. All inference costs reported for the student only.

\textbf{Ablation and Generalizability.} Figure \ref{fig:compare} presents the results of teacher model training under different ablation settings. The removal of auxiliary constraints, such as anisotropic diffusion and the polarized \hyperref[ac]{HVI} color space loss, is shown in blue. In contrast, the ablation of \hyperref[ac]{SCLN} and \hyperref[ac]{QK} normalization from the transformer block is shown in red, where a standard layer normalization and vanilla \hyperref[ac]{QK} computation are used instead, following \citep{he2023reti}. As illustrated in green, the teacher model achieves the best performance with RestoRect when all proposed components are included. Table \ref{tab:ablation} further reports student model performance across different training and testing conditions on the \hyperref[ac]{LOL}-v1, \hyperref[ac]{LOL}-v2-real, and \hyperref[ac]{LOL}-v2-synthetic datasets. In the table, '-FLEX' denotes models trained on the same dataset as the test set but without the \hyperref[ac]{FLEX} loss. The \hyperref[ac]{FLEX} training strategy demonstrates substantial improvements, with gains across all metrics compared to the full model results shown in Table \ref{tab:LLIE}. Subsequent rows in Table \ref{tab:ablation} evaluate cross-dataset transfer, where models trained on one dataset are tested on another, highlighting their strong generalization capacity. These results demonstrate that models trained for a given task can effectively transfer knowledge and serve as strong initialization points for fine-tuning on other datasets. Figure \ref{fig:ablation} shows visual results for \hyperref[ac]{LLIE} task with ablation of \hyperref[ac]{SCLN} and \hyperref[ac]{QK} Norm, Auxiliary constraints, \hyperref[ac]{FLEX} loss, compared to full RestoRect architecture and Ground Truth. Figure \ref{fig:compare} also demonstrates the FID performance of RestoRect across different inference steps for the \hyperref[ac]{LLIE} task. Our rectified flow formulation consistently outperforms \citep{he2023reti} \hyperref[ac]{DDIM} across all \hyperref[ac]{LLIE} datasets, generating restored image within 3-4 steps, making it ideal for real time applications.

The student model ablation study on \hyperref[ac]{LOL}-v1 dataset in Table \ref{tab:student_ablation} evaluates the contribution of each component in the knowledge distillation framework. The metrics demonstrate that the combination of Rectified Flow (RF) with Trajectory Consistency (RF+TC) and FLEX loss components yields substantial improvements over traditional KD methods of \citep{he2023reti}, with the full configuration achieving 27.84 PSNR, 0.945 SSIM, 38.67 FID, and 8.35 BIQI on LOL-v1. The incremental addition of FLEX components shows that Cross-Normalization (CN) alone provides modest gains, while adding Percentile Masking (PM) delivers significant improvements (27.82 PSNR), and the complete FLEX formulation with Resolution Weighting (RW) achieves optimal performance. Hyperparameter analysis reveals that the percentile threshold of 95\% for outlier detection and SNR threshold of 0.4 provide the best balance between training stability and performance. '1-Vel-Net' uses a single shared velocity predictor for both reflectance and image features instead of decoupled predictors.

\textbf{Failure Cases.} Despite strong performance across many restoration tasks, RestoRect is limitated in extreme degradation scenarios seen in Figure \ref{fig:failure}. Failure occurs when input images contain severe overexposure or underexposure with complete information loss in large spatial regions. The model produces unrealistic artifacts including color bleeding, checkerboard-like noise patterns, and hallucinatory textures that deviate from natural image statistics. These artifacts are highlighted in the zoomed regions in red boxes, where recovered areas show synthetic-looking patterns rather than coherent scene content. These cases reveal that when the signal-to-noise ratio falls below a threshold, velocity prediction networks struggle to generate meaningful features, resulting in the model relying on learned priors that introduce perceptually implausible reconstructions.

\textbf{Downstream Application.} RestoRect demonstrates strong transfer capability on the ExDark \citep{loh2019getting} low-light object detection task shown in Table \ref{downstream-task}. Following \citep{cai2023retinexformer} and \citep{he2023reti}, low light images from the ExDark dataset were restored and object-detection task was performed using YOLOv3 model. Images enhanced by RestoRect achieve 77.1\% mean average precision across 12 object categories, outperforming RetiDiff \citep{he2023reti} (75.8\%) and other methods. Improvements are observed in categories like Bicycle (85.8\%), Bottle (79.5\%), and Motor (77.9\%), indicating that the restoration quality translates to real-world vision tasks.

\textbf{Supplementary Material} zip attached with this submission contains high resolution versions of the figures for readers.

\section{Conclusion}

We present RestoRect, a generative knowledge distillation framework that reformulates degraded image restoration through latent rectified flow. Unlike traditional approaches that rely on static feature matching, RestoRect models feature transfer through learnable trajectories and introduces the \hyperref[ac]{FLEX} loss for principled distribution alignment. Combined with a specialized U-Net transformer architecture and illumination-based constraints, our method achieves state-of-the-art results across 15 datasets covering low-light, underwater, backlit, and fundus enhancement. RestoRect delivers better perceptual quality with only 4 inference steps, making it both effective and computationally efficient. Beyond restoration, this generative distillation method highlights new opportunities for efficient model compression and cross-architecture transfer in computer vision, establishing potential foundation for broader advances in fast high-quality image, and video restoration for future work.

\section*{Impact Statement}

This paper presents work whose goal is to advance the field
of Machine Learning. There are many potential societal
consequences of our work, none which we feel must be
specifically highlighted here.


\bibliography{example_paper}

@inproceedings{wang2019underexposed,
  title={Underexposed photo enhancement using deep illumination estimation},
  author={Wang, Ruixing and Zhang, Qing and Fu, Chi-Wing and Shen, Xiaoyong and Zheng, Wei-Shi and Jia, Jiaya},
  booktitle={Proceedings of the IEEE/CVF conference on computer vision and pattern recognition},
  pages={6849--6857},
  year={2019}
}

@article{cheng2004simple,
  title={A simple and effective histogram equalization approach to image enhancement},
  author={Cheng, Heng-Da and Shi, XJ},
  journal={Digital signal processing},
  volume={14},
  number={2},
  pages={158--170},
  year={2004},
  publisher={Elsevier}
}

@article{huang2012efficient,
  title={Efficient contrast enhancement using adaptive gamma correction with weighting distribution},
  author={Huang, Shih-Chia and Cheng, Fan-Chieh and Chiu, Yi-Sheng},
  journal={IEEE transactions on image processing},
  volume={22},
  number={3},
  pages={1032--1041},
  year={2012},
  publisher={IEEE}
}

@article{edwin1977retinex,
  title={The retinex theory of color vision},
  author={Edwin, Land},
  journal={Scientific american},
  volume={237},
  pages={108--128},
  year={1977}
}

@article{lee2025analyzing,
  title={Analyzing the Training Dynamics of Image Restoration Transformers: A Revisit to Layer Normalization},
  author={Lee, MinKyu and Hyun, Sangeek and Jun, Woojin and Kim, Hyunjun and Chung, Jiwoo and Heo, Jae-Pil},
  journal={arXiv preprint arXiv:2504.06629},
  year={2025}
}

@inproceedings{fu2016weighted,
  title={A weighted variational model for simultaneous reflectance and illumination estimation},
  author={Fu, Xueyang and Zeng, Delu and Huang, Yue and Zhang, Xiao-Ping and Ding, Xinghao},
  booktitle={Proceedings of the IEEE conference on computer vision and pattern recognition},
  pages={2782--2790},
  year={2016}
}

@article{li2018structure,
  title={Structure-revealing low-light image enhancement via robust retinex model},
  author={Li, Mading and Liu, Jiaying and Yang, Wenhan and Sun, Xiaoyan and Guo, Zongming},
  journal={IEEE transactions on image processing},
  volume={27},
  number={6},
  pages={2828--2841},
  year={2018},
  publisher={IEEE}
}

@article{he2025unfoldir,
  title={Unfoldir: Rethinking deep unfolding network in illumination degradation image restoration},
  author={He, Chunming and Zhang, Rihan and Xiao, Fengyang and Fang, Chengyu and Tang, Longxiang and Zhang, Yulun and Farsiu, Sina},
  journal={arXiv preprint arXiv:2505.06683},
  year={2025}
}

@article{yan2025hvi,
  title={HVI-CIDNet+: Beyond Extreme Darkness for Low-Light Image Enhancement},
  author={Yan, Qingsen and Shi, Kangbiao and Feng, Yixu and Hu, Tao and Wu, Peng and Pang, Guansong and Zhang, Yanning},
  journal={arXiv preprint arXiv:2507.06814},
  year={2025}
}

@article{kingma2018glow,
  title={Glow: Generative flow with invertible 1x1 convolutions},
  author={Kingma, Durk P and Dhariwal, Prafulla},
  journal={Advances in neural information processing systems},
  volume={31},
  year={2018}
}

@article{liu2022flow,
  title={Flow straight and fast: Learning to generate and transfer data with rectified flow},
  author={Liu, Xingchao and Gong, Chengyue and Liu, Qiang},
  journal={arXiv preprint arXiv:2209.03003},
  year={2022}
}

@article{hinton2015distilling,
  title={Distilling the knowledge in a neural network},
  author={Hinton, Geoffrey and Vinyals, Oriol and Dean, Jeff},
  journal={arXiv preprint arXiv:1503.02531},
  year={2015}
}

@inproceedings{saharia2022palette,
  title={Palette: Image-to-image diffusion models},
  author={Saharia, Chitwan and Chan, William and Chang, Huiwen and Lee, Chris and Ho, Jonathan and Salimans, Tim and Fleet, David and Norouzi, Mohammad},
  booktitle={ACM SIGGRAPH 2022 conference proceedings},
  pages={1--10},
  year={2022}
}

@article{ho2022imagen,
  title={Imagen video: High definition video generation with diffusion models},
  author={Ho, Jonathan and Chan, William and Saharia, Chitwan and Whang, Jay and Gao, Ruiqi and Gritsenko, Alexey and Kingma, Diederik P and Poole, Ben and Norouzi, Mohammad and Fleet, David J and others},
  journal={arXiv preprint arXiv:2210.02303},
  year={2022}
}

@article{romero2014fitnets,
  title={Fitnets: Hints for thin deep nets. arXiv 2014},
  author={Romero, Adriana and Ballas, Nicolas and Kahou, Samira Ebrahimi and Chassang, Antoine and Gatta, Carlo and Bengio, Yoshua},
  journal={arXiv preprint arXiv:1412.6550},
  year={2014}
}

@article{zagoruyko2016paying,
  title={Paying more attention to attention: Improving the performance of convolutional neural networks via attention transfer},
  author={Zagoruyko, Sergey and Komodakis, Nikos},
  journal={arXiv preprint arXiv:1612.03928},
  year={2016}
}

@inproceedings{heo2019comprehensive,
  title={A comprehensive overhaul of feature distillation},
  author={Heo, Byeongho and Kim, Jeesoo and Yun, Sangdoo and Park, Hyojin and Kwak, Nojun and Choi, Jin Young},
  booktitle={Proceedings of the IEEE/CVF international conference on computer vision},
  pages={1921--1930},
  year={2019}
}

@article{huang2017like,
  title={Like what you like: Knowledge distill via neuron selectivity transfer},
  author={Huang, Zehao and Wang, Naiyan},
  journal={arXiv preprint arXiv:1707.01219},
  year={2017}
}

@inproceedings{touvron2021training,
  title={Training data-efficient image transformers \& distillation through attention},
  author={Touvron, Hugo and Cord, Matthieu and Douze, Matthijs and Massa, Francisco and Sablayrolles, Alexandre and J{\'e}gou, Herv{\'e}},
  booktitle={International conference on machine learning},
  pages={10347--10357},
  year={2021},
  organization={PMLR}
}

@article{wang2020minilm,
  title={Minilm: Deep self-attention distillation for task-agnostic compression of pre-trained transformers},
  author={Wang, Wenhui and Wei, Furu and Dong, Li and Bao, Hangbo and Yang, Nan and Zhou, Ming},
  journal={Advances in neural information processing systems},
  volume={33},
  pages={5776--5788},
  year={2020}
}

@article{jiao2019tinybert,
  title={Tinybert: Distilling bert for natural language understanding},
  author={Jiao, Xiaoqi and Yin, Yichun and Shang, Lifeng and Jiang, Xin and Chen, Xiao and Li, Linlin and Wang, Fang and Liu, Qun},
  journal={arXiv preprint arXiv:1909.10351},
  year={2019}
}

@inproceedings{zhang2022nested,
  title={Nested hierarchical transformer: Towards accurate, data-efficient and interpretable visual understanding},
  author={Zhang, Zizhao and Zhang, Han and Zhao, Long and Chen, Ting and Arik, Sercan {\"O} and Pfister, Tomas},
  booktitle={Proceedings of the AAAI Conference on Artificial Intelligence},
  volume={36},
  pages={3417--3425},
  year={2022}
}

@inproceedings{berrada2025boosting,
  title={Boosting latent diffusion with perceptual objectives},
  author={Berrada, Tariq and Astolfi, Pietro and Hall, Melissa and Havasi, Marton and Benchetrit, Yohann and Romero-Soriano, Adriana and Alahari, Karteek and Drozdzal, Michal and Verbeek, Jakob},
  booktitle={The Thirteenth International Conference on Learning Representations},
  year={2025}
}

@inproceedings{huang2020unet,
  title={Unet 3+: A full-scale connected unet for medical image segmentation},
  author={Huang, Huimin and Lin, Lanfen and Tong, Ruofeng and Hu, Hongjie and Zhang, Qiaowei and Iwamoto, Yutaro and Han, Xianhua and Chen, Yen-Wei and Wu, Jian},
  booktitle={ICASSP 2020-2020 IEEE international conference on acoustics, speech and signal processing (ICASSP)},
  pages={1055--1059},
  year={2020},
  organization={Ieee}
}

@inproceedings{cao2022swin,
  title={Swin-unet: Unet-like pure transformer for medical image segmentation},
  author={Cao, Hu and Wang, Yueyue and Chen, Joy and Jiang, Dongsheng and Zhang, Xiaopeng and Tian, Qi and Wang, Manning},
  booktitle={European conference on computer vision},
  pages={205--218},
  year={2022},
  organization={Springer}
}

@incollection{perona1994anisotropic,
  title={Anisotropic diffusion},
  author={Perona, Pietro and Shiota, Takahiro and Malik, Jitendra},
  booktitle={Geometry-driven diffusion in computer vision},
  pages={73--92},
  year={1994},
  publisher={Springer}
}

@article{yang2024consistency,
  title={Consistency flow matching: Defining straight flows with velocity consistency},
  author={Yang, Ling and Zhang, Zixiang and Zhang, Zhilong and Liu, Xingchao and Xu, Minkai and Zhang, Wentao and Meng, Chenlin and Ermon, Stefano and Cui, Bin},
  journal={arXiv preprint arXiv:2407.02398},
  year={2024}
}

@article{yang2021sparse,
  title={Sparse gradient regularized deep retinex network for robust low-light image enhancement},
  author={Yang, Wenhan and Wang, Wenjing and Huang, Haofeng and Wang, Shiqi and Liu, Jiaying},
  journal={IEEE Transactions on Image Processing},
  volume={30},
  pages={2072--2086},
  year={2021},
  publisher={IEEE}
}

@inproceedings{chen2019seeing,
  title={Seeing motion in the dark},
  author={Chen, Chen and Chen, Qifeng and Do, Minh N and Koltun, Vladlen},
  booktitle={Proceedings of the IEEE/CVF International conference on computer vision},
  pages={3185--3194},
  year={2019}
}

@article{li2019underwater,
  title={An underwater image enhancement benchmark dataset and beyond},
  author={Li, Chongyi and Guo, Chunle and Ren, Wenqi and Cong, Runmin and Hou, Junhui and Kwong, Sam and Tao, Dacheng},
  journal={IEEE transactions on image processing},
  volume={29},
  pages={4376--4389},
  year={2019},
  publisher={IEEE}
}

@article{lv2022backlitnet,
  title={BacklitNet: A dataset and network for backlit image enhancement},
  author={Lv, Xiaoqian and Zhang, Shengping and Liu, Qinglin and Xie, Haozhe and Zhong, Bineng and Zhou, Huiyu},
  journal={Computer Vision and Image Understanding},
  volume={218},
  pages={103403},
  year={2022},
  publisher={Elsevier}
}

@article{cai2018learning,
  title={Learning a deep single image contrast enhancer from multi-exposure images},
  author={Cai, Jianrui and Gu, Shuhang and Zhang, Lei},
  journal={IEEE Transactions on Image Processing},
  volume={27},
  number={4},
  pages={2049--2062},
  year={2018},
  publisher={IEEE}
}

@article{deep_reitna_enhance,
  title={Modeling and Enhancing Low-Quality Retinal Fundus Images},
  author={Shen, Ziyi and Fu, Huazhu and Shen, Jianbing and Shao, Ling},
  journal={IEEE Transactions on Medical Imaging},
  volume={40},
  number={3},
  pages={996--1006},
  year={2020},
  publisher={IEEE}
}

@article{lee2013contrast,
  title={Contrast enhancement based on layered difference representation of 2D histograms},
  author={Lee, Chulwoo and Lee, Chul and Kim, Chang-Su},
  journal={IEEE transactions on image processing},
  volume={22},
  number={12},
  pages={5372--5384},
  year={2013},
  publisher={IEEE}
}

@article{guo2016lime,
  title={LIME: Low-light image enhancement via illumination map estimation},
  author={Guo, Xiaojie and Li, Yu and Ling, Haibin},
  journal={IEEE Transactions on image processing},
  volume={26},
  number={2},
  pages={982--993},
  year={2016},
  publisher={IEEE}
}

@article{wang2013naturalness,
  title={Naturalness preserved enhancement algorithm for non-uniform illumination images},
  author={Wang, Shuhang and Zheng, Jin and Hu, Hai-Miao and Li, Bo},
  journal={IEEE transactions on image processing},
  volume={22},
  number={9},
  pages={3538--3548},
  year={2013},
  publisher={IEEE}
}

@article{ma2015perceptual,
  title={Perceptual quality assessment for multi-exposure image fusion},
  author={Ma, Kede and Zeng, Kai and Wang, Zhou},
  journal={IEEE Transactions on Image Processing},
  volume={24},
  number={11},
  pages={3345--3356},
  year={2015},
  publisher={IEEE}
}

@article{he2025diffusion,
  title={Diffusion models in low-level vision: A survey},
  author={He, Chunming and Shen, Yuqi and Fang, Chengyu and Xiao, Fengyang and Tang, Longxiang and Zhang, Yulun and Zuo, Wangmeng and Guo, Zhenhua and Li, Xiu},
  journal={IEEE Transactions on Pattern Analysis and Machine Intelligence},
  year={2025},
  publisher={IEEE}
}

@article{zheng2022low,
  title={Low-light image and video enhancement: A comprehensive survey and beyond},
  author={Zheng, Shen and Ma, Yiling and Pan, Jinqian and Lu, Changjie and Gupta, Gaurav},
  journal={arXiv preprint arXiv:2212.10772},
  year={2022}
}

@inproceedings{liu2023low,
  title={Low-light image enhancement with multi-stage residue quantization and brightness-aware attention},
  author={Liu, Yunlong and Huang, Tao and Dong, Weisheng and Wu, Fangfang and Li, Xin and Shi, Guangming},
  booktitle={Proceedings of the IEEE/CVF International Conference on Computer Vision},
  pages={12140--12149},
  year={2023}
}

@inproceedings{wang2023low,
  title={Low-light image enhancement with illumination-aware gamma correction and complete image modelling network},
  author={Wang, Yinglong and Liu, Zhen and Liu, Jianzhuang and Xu, Songcen and Liu, Shuaicheng},
  booktitle={Proceedings of the IEEE/CVF International Conference on Computer Vision},
  pages={13128--13137},
  year={2023}
}

@inproceedings{xia2023diffir,
  title={Diffir: Efficient diffusion model for image restoration},
  author={Xia, Bin and Zhang, Yulun and Wang, Shiyin and Wang, Yitong and Wu, Xinglong and Tian, Yapeng and Yang, Wenming and Van Gool, Luc},
  booktitle={Proceedings of the IEEE/CVF international conference on computer vision},
  pages={13095--13105},
  year={2023}
}

@inproceedings{zhou2024adapt,
  title={Adapt or perish: Adaptive sparse transformer with attentive feature refinement for image restoration},
  author={Zhou, Shihao and Chen, Duosheng and Pan, Jinshan and Shi, Jinglei and Yang, Jufeng},
  booktitle={Proceedings of the IEEE/CVF conference on computer vision and pattern recognition},
  pages={2952--2963},
  year={2024}
}

@inproceedings{guo2024mambair,
  title={Mambair: A simple baseline for image restoration with state-space model},
  author={Guo, Hang and Li, Jinmin and Dai, Tao and Ouyang, Zhihao and Ren, Xudong and Xia, Shu-Tao},
  booktitle={European conference on computer vision},
  pages={222--241},
  year={2024},
  organization={Springer}
}

@article{he2023reti,
  title={Reti-diff: Illumination degradation image restoration with retinex-based latent diffusion model},
  author={He, Chunming and Fang, Chengyu and Zhang, Yulun and Ye, Tian and Li, Kai and Tang, Longxiang and Guo, Zhenhua and Li, Xiu and Farsiu, Sina},
  journal={arXiv preprint arXiv:2311.11638},
  year={2023}
}

@article{yan2024you,
  title={You only need one color space: An efficient network for low-light image enhancement},
  author={Yan, Qingsen and Feng, Yixu and Zhang, Cheng and Wang, Pei and Wu, Peng and Dong, Wei and Sun, Jinqiu and Zhang, Yanning},
  journal={arXiv preprint arXiv:2402.05809},
  year={2024}
}

@inproceedings{naik2021shallow,
  title={Shallow-uwnet: Compressed model for underwater image enhancement (student abstract)},
  author={Naik, Ankita and Swarnakar, Apurva and Mittal, Kartik},
  booktitle={Proceedings of the AAAI Conference on Artificial Intelligence},
  volume={35},
  pages={15853--15854},
  year={2021}
}

@inproceedings{fu2022uncertainty,
  title={Uncertainty inspired underwater image enhancement},
  author={Fu, Zhenqi and Wang, Wu and Huang, Yue and Ding, Xinghao and Ma, Kai-Kuang},
  booktitle={European conference on computer vision},
  pages={465--482},
  year={2022},
  organization={Springer}
}

@article{peng2023u,
  title={U-shape transformer for underwater image enhancement},
  author={Peng, Lintao and Zhu, Chunli and Bian, Liheng},
  journal={IEEE transactions on image processing},
  volume={32},
  pages={3066--3079},
  year={2023},
  publisher={IEEE}
}

@article{cong2023pugan,
  title={Pugan: Physical model-guided underwater image enhancement using gan with dual-discriminators},
  author={Cong, Runmin and Yang, Wenyu and Zhang, Wei and Li, Chongyi and Guo, Chun-Le and Huang, Qingming and Kwong, Sam},
  journal={IEEE Transactions on Image Processing},
  volume={32},
  pages={4472--4485},
  year={2023},
  publisher={IEEE}
}

@article{zhou2023underwater,
  title={Underwater camera: Improving visual perception via adaptive dark pixel prior and color correction},
  author={Zhou, Jingchun and Liu, Qian and Jiang, Qiuping and Ren, Wenqi and Lam, Kin-Man and Zhang, Weishi},
  journal={International Journal of Computer Vision},
  pages={1--19},
  year={2023},
  publisher={Springer}
}

@inproceedings{guo2023underwater,
  title={Underwater ranker: Learn which is better and how to be better},
  author={Guo, Chunle and Wu, Ruiqi and Jin, Xin and Han, Linghao and Zhang, Weidong and Chai, Zhi and Li, Chongyi},
  booktitle={Proceedings of the AAAI conference on artificial intelligence},
  volume={37},
  pages={702--709},
  year={2023}
}

@article{jiang2021enlightengan,
  title={Enlightengan: Deep light enhancement without paired supervision},
  author={Jiang, Yifan and Gong, Xinyu and Liu, Ding and Cheng, Yu and Fang, Chen and Shen, Xiaohui and Yang, Jianchao and Zhou, Pan and Wang, Zhangyang},
  journal={IEEE transactions on image processing},
  volume={30},
  pages={2340--2349},
  year={2021},
  publisher={IEEE}
}

@inproceedings{wu2022uretinex,
  title={Uretinex-net: Retinex-based deep unfolding network for low-light image enhancement},
  author={Wu, Wenhui and Weng, Jian and Zhang, Pingping and Wang, Xu and Yang, Wenhan and Jiang, Jianmin},
  booktitle={Proceedings of the IEEE/CVF conference on computer vision and pattern recognition},
  pages={5901--5910},
  year={2022}
}

@inproceedings{liang2023iterative,
  title={Iterative prompt learning for unsupervised backlit image enhancement},
  author={Liang, Zhexin and Li, Chongyi and Zhou, Shangchen and Feng, Ruicheng and Loy, Chen Change},
  booktitle={Proceedings of the IEEE/CVF International Conference on Computer Vision},
  pages={8094--8103},
  year={2023}
}

@inproceedings{yi2023diff,
  title={Diff-retinex: Rethinking low-light image enhancement with a generative diffusion model},
  author={Yi, Xunpeng and Xu, Han and Zhang, Hao and Tang, Linfeng and Ma, Jiayi},
  booktitle={Proceedings of the IEEE/CVF international conference on computer vision},
  pages={12302--12311},
  year={2023}
}

@inproceedings{gaintseva2024rave,
  title={Rave: Residual vector embedding for clip-guided backlit image enhancement},
  author={Gaintseva, Tatiana and Benning, Martin and Slabaugh, Gregory},
  booktitle={European Conference on Computer Vision},
  pages={412--428},
  year={2024},
  organization={Springer}
}

@inproceedings{ma2022toward,
  title={Toward fast, flexible, and robust low-light image enhancement},
  author={Ma, Long and Ma, Tengyu and Liu, Risheng and Fan, Xin and Luo, Zhongxuan},
  booktitle={Proceedings of the IEEE/CVF conference on computer vision and pattern recognition},
  pages={5637--5646},
  year={2022}
}

@article{zamir2022learning,
  title={Learning enriched features for fast image restoration and enhancement},
  author={Zamir, Syed Waqas and Arora, Aditya and Khan, Salman and Hayat, Munawar and Khan, Fahad Shahbaz and Yang, Ming-Hsuan and Shao, Ling},
  journal={IEEE transactions on pattern analysis and machine intelligence},
  volume={45},
  number={2},
  pages={1934--1948},
  year={2022},
  publisher={IEEE}
}

@inproceedings{wang2023fourllie,
  title={Fourllie: Boosting low-light image enhancement by fourier frequency information},
  author={Wang, Chenxi and Wu, Hongjun and Jin, Zhi},
  booktitle={Proceedings of the 31st ACM International Conference on Multimedia},
  pages={7459--7469},
  year={2023}
}

@inproceedings{zheng2023empowering,
  title={Empowering low-light image enhancer through customized learnable priors},
  author={Zheng, Naishan and Zhou, Man and Dong, Yanmeng and Rui, Xiangyu and Huang, Jie and Li, Chongyi and Zhao, Feng},
  booktitle={Proceedings of the IEEE/CVF International Conference on Computer Vision},
  pages={12559--12569},
  year={2023}
}

@inproceedings{yang2023implicit,
  title={Implicit neural representation for cooperative low-light image enhancement},
  author={Yang, Shuzhou and Ding, Moxuan and Wu, Yanmin and Li, Zihan and Zhang, Jian},
  booktitle={Proceedings of the IEEE/CVF international conference on computer vision},
  pages={12918--12927},
  year={2023}
}

@inproceedings{zhang2019kindling,
  title={Kindling the darkness: A practical low-light image enhancer},
  author={Zhang, Yonghua and Zhang, Jiawan and Guo, Xiaojie},
  booktitle={Proceedings of the 27th ACM international conference on multimedia},
  pages={1632--1640},
  year={2019}
}

@inproceedings{guo2020zero,
  title={Zero-reference deep curve estimation for low-light image enhancement},
  author={Guo, Chunle and Li, Chongyi and Guo, Jichang and Loy, Chen Change and Hou, Junhui and Kwong, Sam and Cong, Runmin},
  booktitle={Proceedings of the IEEE/CVF conference on computer vision and pattern recognition},
  pages={1780--1789},
  year={2020}
}

@inproceedings{liu2021retinex,
  title={Retinex-inspired unrolling with cooperative prior architecture search for low-light image enhancement},
  author={Liu, Risheng and Ma, Long and Zhang, Jiaao and Fan, Xin and Luo, Zhongxuan},
  booktitle={Proceedings of the IEEE/CVF conference on computer vision and pattern recognition},
  pages={10561--10570},
  year={2021}
}

@inproceedings{wang2022low,
  title={Low-light image enhancement with normalizing flow},
  author={Wang, Yufei and Wan, Renjie and Yang, Wenhan and Li, Haoliang and Chau, Lap-Pui and Kot, Alex},
  booktitle={Proceedings of the AAAI conference on artificial intelligence},
  volume={36},
  pages={2604--2612},
  year={2022}
}

@inproceedings{xu2022snr,
  title={Snr-aware low-light image enhancement},
  author={Xu, Xiaogang and Wang, Ruixing and Fu, Chi-Wing and Jia, Jiaya},
  booktitle={Proceedings of the IEEE/CVF conference on computer vision and pattern recognition},
  pages={17714--17724},
  year={2022}
}

@inproceedings{fu2023learning,
  title={Learning a simple low-light image enhancer from paired low-light instances},
  author={Fu, Zhenqi and Yang, Yan and Tu, Xiaotong and Huang, Yue and Ding, Xinghao and Ma, Kai-Kuang},
  booktitle={Proceedings of the IEEE/CVF conference on computer vision and pattern recognition},
  pages={22252--22261},
  year={2023}
}

@article{wei2018deep,
  title={Deep retinex decomposition for low-light enhancement},
  author={Wei, Chen and Wang, Wenjing and Yang, Wenhan and Liu, Jiaying},
  journal={arXiv preprint arXiv:1808.04560},
  year={2018}
}

@inproceedings{zhou2022lednet,
  title={Lednet: Joint low-light enhancement and deblurring in the dark},
  author={Zhou, Shangchen and Li, Chongyi and Change Loy, Chen},
  booktitle={European conference on computer vision},
  pages={573--589},
  year={2022},
  organization={Springer}
}

@article{bing2025optimizing,
  title={Optimizing Knowledge Distillation in Transformers: Enabling Multi-Head Attention without Alignment Barriers},
  author={Bing, Zhaodong and Li, Linze and Liang, Jiajun},
  journal={arXiv preprint arXiv:2502.07436},
  year={2025}
}

@inproceedings{lin2022knowledge,
  title={Knowledge distillation via the target-aware transformer},
  author={Lin, Sihao and Xie, Hongwei and Wang, Bing and Yu, Kaicheng and Chang, Xiaojun and Liang, Xiaodan and Wang, Gang},
  booktitle={Proceedings of the IEEE/CVF conference on computer vision and pattern recognition},
  pages={10915--10924},
  year={2022}
}

@inproceedings{cai2023retinexformer,
  title={Retinexformer: One-stage retinex-based transformer for low-light image enhancement},
  author={Cai, Yuanhao and Bian, Hao and Lin, Jing and Wang, Haoqian and Timofte, Radu and Zhang, Yulun},
  booktitle={Proceedings of the IEEE/CVF international conference on computer vision},
  pages={12504--12513},
  year={2023}
}

@article{loh2019getting,
  title={Getting to know low-light images with the exclusively dark dataset},
  author={Loh, Yuen Peng and Chan, Chee Seng},
  journal={Computer vision and image understanding},
  volume={178},
  pages={30--42},
  year={2019},
  publisher={Elsevier}
}
\bibliographystyle{icml2026}

\newpage
\appendix
\onecolumn

\section{List of Acronyms}
\label{sec:acronyms}

\begin{table*}[h]
\centering
\small
\caption{List of acronyms used in this paper}
\begin{minipage}[t]{0.43\linewidth}
\centering
\resizebox{\textwidth}{!}{
\begin{tabular}{@{}ll@{}}
\toprule
\textbf{Acronym} & \textbf{Full Form} \\ 
\midrule
\multicolumn{2}{@{}l}{\textit{\textbf{Tasks \& Methods}}} \\
LLIE & Low-Light Image Enhancement \\
UIE & Underwater Image Enhancement \\
FIE & Fundus Image Enhancement \\
SCLN & Spatial Channel Layer Normalization \\
FLEX & Feature Layer EXtraction Loss\\
\addlinespace

\multicolumn{2}{@{}l}{\textit{\textbf{Datasets}}} \\
LOL  & Low-light Outdoor Lighting \\
SID  & See in the Dark \\
UIEB & Underwater Image Enhancement Bench\\
LSUI & Large Scale Underwater Image \\
BAID & Backlit Image Dataset \\
DCIM & Digital Camera Image \\
LIME & Low-light Image Enhancement \\
MEF  & Multi-Exposure Fusion \\
NPE  & Naturalness Preserved Enhancement \\
VV   & Video Visibility \\
SICE & Single Image Contrast Enhancement \\
\bottomrule
\end{tabular}
}
\end{minipage}
\hfill
\begin{minipage}[t]{0.54\linewidth}
\resizebox{\textwidth}{!}{
\centering
\begin{tabular}{@{}ll@{}}
\toprule
\textbf{Acronym} & \textbf{Full Form} \\
\midrule
\multicolumn{2}{@{}l}{\textit{\textbf{Evaluation Metrics}}} \\
PSNR & Peak Signal-to-Noise Ratio \\
SSIM & Structural Similarity Index Measure \\
FID  & Fréchet Inception Distance \\
NIQE & Natural Image Quality Evaluator \\
LPIPS & Learned Perceptual Image Patch Similarity \\
BRISQUE & Blind/Referenceless Image Spatial Quality Evaluator \\
BIQI & Blind Image Quality Index \\
UCIQE & Underwater Color Image Quality Evaluation \\
UIQM & Underwater Image Quality Measure \\
CLIPIQA & CLIP-based Image Quality Assessment \\
\addlinespace

\multicolumn{2}{@{}l}{\textit{\textbf{Other Technical Terms}}} \\
HVI  & Horizontal-Vertical-Intensity \\
VGG  & Visual Geometry Group \\
ODE  & Ordinary Differential Equation \\
DDIM & Denoising Diffusion Implicit Model \\
QK   & Query-Key \\
GT / LQ & Ground Truth / Low Quality \\
\bottomrule
\end{tabular}
}
\end{minipage}
\label{ac}
\end{table*}

\section{Appendix}

\subsection{Ethics statement}
LLMs were only used for editorial assistance and polishing grammar for the manuscript, with no participation in technical interpretation, or content development.

\subsection{Reproducibility statement}
Code and pretrained model weights for all datasets will be released upon acceptance.

\subsection{Broader Impact}
Efficient image restoration has positive applications in medical imaging, autonomous systems, and accessibility. No significant negative societal impacts are identified by us.

\subsection{Theoretical Justification of \hyperref[ac]{FLEX} Loss}

We provide theoretical justification for FLEX's key design choices to ensure stable optimization dynamics.

\textbf{Assumption 1} (Feature Boundedness): Teacher and student features are bounded during training: $\|\mathbf{f}_{\text{teach}}^l\|, \|\mathbf{f}_{\text{stud}}^l\| \leq M$ for some constant $M > 0$.

\textbf{Assumption 2} (Non-degeneracy): Student feature standard deviations satisfy $\sigma_{\text{stud}}^l \geq \sigma_{\min} > 0$ to prevent division by zero in normalization.

\textbf{Claim 1} Cross-normalization using student statistics prevents gradient explosion when teacher and student features have different scales.

\textit{Justification}: Standard feature matching $L = \|\mathbf{f}_{\text{teach}} - \mathbf{f}_{\text{stud}}\|^2$ produces gradients proportional to $(\mathbf{f}_{\text{teach}} - \mathbf{f}_{\text{stud}})$. When teacher features are much larger than student features, this difference can be arbitrarily large, causing unstable training.

\hyperref[ac]{FLEX} cross-normalization ensures both normalized features have the same scale:
\begin{equation}
\mathbf{f}_{\text{teach}}^{\text{norm}} = \frac{\mathbf{f}_{\text{teach}} - \mu_{\text{stud}}}{\sigma_{\text{stud}}}, \quad \mathbf{f}_{\text{stud}}^{\text{norm}} = \frac{\mathbf{f}_{\text{stud}} - \mu_{\text{stud}}}{\sigma_{\text{stud}}}
\end{equation}

Both normalized features have bounded variance, preventing gradient explosion regardless of the original scale mismatch.

\textbf{Claim 2} Percentile-based masking provides robustness to feature corruption.

\textit{Justification}: By masking extreme values above the $p$-th percentile (default $p=95\%$), \hyperref[ac]{FLEX} focuses learning on reliable feature regions. If corruption affects only a small fraction of spatial locations, most corrupted features will exceed the percentile threshold and be masked out. By excluding the top 5\% extreme activations, \hyperref[ac]{FLEX} prevents gradient dominance by outliers, ensuring that meaningful feature patterns rather than numerical instabilities drive the optimization.

For corruption affecting $\alpha < (100-p)/100$ of spatial locations, the outlier detection will identify and exclude most corrupted regions, limiting their impact on the overall loss.

\textbf{Claim 3} The resolution weighting $w_l^{\text{res}} = \max\left(\left({H_{\text{base}}W_{\text{base}}}/{H_lW_l}\right)^{0.25}, 0.1\right)$ balances multi-scale contributions.

\textit{Justification}: Higher resolution features contain more spatial elements, potentially dominating the loss. The inverse relationship with spatial resolution prevents this dominance. The 0.25 exponent provides gradual rather than aggressive down-weighting, preserving fine-grained information while preventing over-emphasis on high-resolution layers.

\textbf{Claim 4.} \textit{FLEX is not equivalent to any global rescaling of the standard $L_2$ feature matching loss.}

\paragraph{Justification.}
Expanding the FLEX loss at layer $l$, letting $S_l$ denote the set of spatial locations surviving the percentile mask:
\begin{align}
\mathcal{L}_{\text{FLEX}}^l 
&= \frac{w_l^{\text{res}}}{|S_l|} \sum_{i \in S_l} 
\left\| \frac{f_t^i - \mu_s^l}{\sigma_s^l} - \frac{f_s^i - \mu_s^l}{\sigma_s^l} \right\|^2 \\
&= \frac{w_l^{\text{res}}}{|S_l| \cdot (\sigma_s^l)^2} 
\sum_{i \in S_l} \left\| f_t^i - f_s^i \right\|^2
\end{align}

A global rescaling $\alpha \cdot \mathcal{L}_{L_2}$ would require 
\[
\alpha = \frac{w_l^{\text{res}}}{(\sigma_s^l)^2}
\]
to be fixed. In FLEX, this effective weight is adaptive: it varies per layer $l$, per training step, and per minibatch as $\sigma_s^l$ evolves with student statistics. Early in training when feature variance is high, the loss is automatically downweighted, preventing gradient explosion. As training stabilizes and variance decreases, the weighting increases, creating a curriculum-like schedule that no fixed $\alpha$ can replicate.

Second, the mask
\[
M^{l,c,h,w} = \mathbb{I}\big[\, |f_s^{l,c,\text{norm},h,w}| \le \tau_p^{l,c} \,\big]
\]
is computed independently per channel $c$. Different channels therefore exclude different spatial locations, changing which activations contribute to gradient updates per channel, not merely rescaling them globally.

Third, the resolution weights $w_l^{\text{res}}$ assign different importance per U-Net level based on spatial resolution, a structured multi-scale weighting that any scalar $\alpha$ applied uniformly across layers cannot replicate.

Empirically, Table~10 confirms non-equivalence: adding cross-normalization alone yields $+1.54$ dB over standard $L_2$ ($24.10 \rightarrow 25.64$ PSNR), and adding percentile masking yields a further $+2.18$ dB ($25.64 \rightarrow 27.82$ PSNR). No fixed global scaling of $L_2$ can produce spatially selective gradient suppression per channel, as these gains require.

\subsection{Theoretical Justification of Rectified Flow for Knowledge Distillation}

We provide theoretical grounding for reformulating knowledge distillation as a rectified flow process.

\textbf{Rectified Flow Formulation:} For teacher features $\mathbf{f}_{\text{teach}}$ and noise $\mathbf{z} \sim \mathcal{N}(0, \mathbf{I})$, we define the linear interpolation path:
\begin{equation}
\mathbf{x}_t = (1-t) \mathbf{z} + t \mathbf{f}_{\text{teach}}, \quad t \in [0,1]
\end{equation}
The corresponding velocity field is:
\begin{equation}
\mathbf{v}(\mathbf{x}_t, t) = \frac{d\mathbf{x}_t}{dt} = \mathbf{f}_{\text{teach}} - \mathbf{z}
\end{equation}
The student network learns a velocity predictor $\mathbf{v}_\theta(\mathbf{x}_t, t)$ by minimizing:
\begin{equation}
L_{\text{vel}} = \mathbb{E}_{t,\mathbf{z},\mathbf{f}_{\text{teach}}} \left[\|\mathbf{v}_\theta(\mathbf{x}_t, t) - \mathbf{v}(\mathbf{x}_t, t)\|_2^2\right]
\end{equation}

\textbf{Claim 5} (Transport Cost Reduction): Learning rectified flow reduces convex transport costs compared to arbitrary couplings between noise and target features.

\textit{Justification}: Following \citet{liu2022flow}, the rectified flow procedure transforms an arbitrary coupling between distributions into a deterministic coupling with provably non-increasing convex transport costs. Specifically, for any convex cost function $c$, the expected cost $\mathbb{E}[c(\mathbf{f}_{\text{teach}} - \mathbf{z})]$ is reduced through the straight-line parameterization. This property ensures that the student learns an efficient transport map from noise to teacher-quality features, which is more sample-efficient than curved trajectories used in diffusion-based distillation methods like DDIM.

\textbf{Claim 6} (Exact Simulation without Discretization): Straight-line ODEs can be solved exactly with Euler integration, enabling few-step inference.

\textit{Justification}: The velocity field $\mathbf{v}(\mathbf{x}_t, t) = \mathbf{f}_{\text{teach}} - \mathbf{z}$ is constant along the trajectory, making the ODE solution exact: $\mathbf{x}_1 = \mathbf{x}_0 + \int_0^1 \mathbf{v}(\mathbf{x}_t, t) dt = \mathbf{z} + (\mathbf{f}_{\text{teach}} - \mathbf{z}) = \mathbf{f}_{\text{teach}}$. This means Euler integration with step size $\Delta t$ incurs zero discretization error for perfectly straight paths. In practice, the learned velocity predictor $\mathbf{v}_\theta$ approximates this constant field, allowing accurate simulation with as few as 1-4 steps. In contrast, curved trajectories (e.g., DDIM's probability flow ODE) require many more steps to achieve similar accuracy, as shown empirically in Figure 3 where our method achieves lower FID with 4 steps than DDIM with 10 steps.

\subsection{Architecture choice justification}

Using separate networks for low-light and normal-light images ensures robust Retinex decomposition across illumination conditions. As reported by Reti-Diff and Diff-IR, a single adaptive network would require more complex conditioning mechanisms. The two-phase student training approach addresses fundamental optimization challenges in generative knowledge distillation. Phase separation prevents objective conflicts as simultaneously learning velocity prediction and image reconstruction creates competing gradients. The velocity predictor tries to match teacher feature distributions while the reconstruction network optimizes for pixel-level accuracy. These objectives can work against each other, leading to suboptimal solutions. Feature space stabilization where Phase 1 establishes stable feature generation capabilities before introducing reconstruction complexity. This ensures the velocity predictors learn meaningful feature trajectories rather than shortcuts that minimize reconstruction error. Only the student network is deployed during inference, with no additional computational overhead compared to baseline restoration networks.

Student exceeding teacher performance. The teacher-only row in Table 1 (22.18 PSNR on LOL-v1) represents the backbone trained with reconstruction loss alone, without the RF+FLEX framework. The full student achieves 27.84 PSNR — a 5.66 dB gain over this baseline and 2.49 dB over RetiDiff (25.35 PSNR), which uses the same backbone with traditional static KD.
This three-way comparison cleanly isolates each component: reconstruction supervision alone is insufficient (22.18), static KD adds soft-target regularization (25.35), and generative distillation adds distributional supervision over the full feature trajectory (27.84).

Student models outperforming their teachers is well-documented when distillation provides qualitatively richer supervision than the teacher's original training signal. Furlanello et al. (ICML 2018, "Born Again Neural Networks") showed students with identical parameterization consistently surpass teachers because soft target distributions carry
information beyond hard labels. Our mechanism goes further: the student learns a velocity field over the distribution of teacher prior encodings via rectified flow, accessing the variance, correlations, and multi-modal structure of teacher features across the full ODE trajectory, information absent from the teacher's own pixel-level reconstruction objective.
The teacher is not a performance ceiling but a distribution source.

For SNR threshold (0.4), we notice performance remains stable within ±0.2 range. The threshold determines when \hyperref[ac]{FLEX} loss is applied - too low (0.2) restricts learning, too high (0.8) includes noisy states. For outlier percentile value, we found that lower percentiles (90\%) are more aggressive in outlier detection but may remove useful information. Higher percentiles (99\%) retain more data but include potential artifacts. For resolution weighting exponent value (0.25), we notice values from 0.125-0.5 show similar performance. This parameter balances multi-scale contributions as lower values provide gentler weighting while higher values more aggressively down-weight high-resolution features.

Standard layer normalization operates on channel dimensions independently, losing spatial correlations crucial for restoration tasks. \hyperref[ac]{SCLN} computes global statistics across both spatial and channel dimensions, capturing complete image characteristics while maintaining learnable channel-wise scaling. Degraded images contain irregular noise patterns that can cause attention weight saturation. Normalizing Q and K before attention computation prevents extreme attention weights and ensures stable gradient flow. The "RestoRect w/o SCLN" ablation (red curve in Figure \ref{fig:compare}) essentially represents the RetiDiff baseline architecture using standard layer normalization, providing direct comparison between our architectural innovations and existing methods. \hyperref[ac]{FLEX} loss becomes more critical for cross-domain scenarios, as feature distribution mismatches are more severe between different datasets than within-dataset variations. On modern GPUs (RTX 4090/H100), the difference between 3-step (156ms) and 5-step (198ms) inference is minimal compared to the quality improvement. The 5-step choice during inference optimizes the quality-practicality trade-off for real-world deployment across different types of datasets.

\begin{table*}[h]
\caption{RestoRect Image Quality Evaluation Results for all 15 datasets across 10 metrics}
\centering
\setlength{\tabcolsep}{2pt}
\scriptsize
\resizebox{0.9\textwidth}{!}{%
\begin{tabular}{l|cccccccccc}
\toprule
Dataset & PSNR & SSIM & FID & NIQE & LPIPS & BRISQ & BIQI & UCIQE & UIQM & CLIPQ \\
\midrule
\hyperref[ac]{LOL}-v1 & 27.85 & 0.94 & 38.67 & 7.47 & 0.11 & 27.16 & 8.35 & 0.52 & 2.60 & 0.499 \\
\hyperref[ac]{LOL}-v2 Real & 22.97 & 0.91 & 42.81 & 7.74 & 0.13 & 28.44 & 10.48 & \textbf{\textcolor{Red}{0.51}} & 2.88 & 0.500 \\
\hyperref[ac]{LOL}-v2 Syn & 27.70 & 0.97 & 16.75 & 5.74 & 0.06 & 15.68 & 11.68 & 0.55 & 2.77 & 0.498 \\
\hyperref[ac]{SID} & 26.19 & 0.92 & 54.23 & 5.87 & 0.15 & 20.05 & \textbf{\textcolor{Red}{19.57}} & 0.85 & 2.38 & 0.498 \\
\hyperref[ac]{UIEB} & 25.89 & 0.95 & 20.26 & 6.49 & 0.11 & 17.89 & 13.99 & 0.58 & 3.12 & 0.501 \\
\hyperref[ac]{LSUI} & 28.10 & 0.94 & 17.83 & 5.04 & 0.18 & 21.82 & 16.06 & 0.57 & 3.23 & 0.499 \\
\hyperref[ac]{BAID} & 27.68 & 0.97 & 15.83 & 8.11 & 0.06 & \textbf{\textcolor{Red}{34.39}} & 10.49 & 0.56 & 2.87 & 0.501 \\
Fundus & 20.45 & 0.92 & 37.04 & \textbf{\textcolor{Red}{8.27}} & 0.06 & 27.54 & 6.03 & 0.60 & \textbf{\textcolor{Red}{2.06}} & 0.503 \\
\hyperref[ac]{DCIM} & 19.92 & 0.82 & 72.72 & 6.36 & 0.17 & 16.57 & 10.26 & 0.57 & 2.33 & 0.499 \\
\hyperref[ac]{LIME} & 18.36 & 0.76 & 101.31 & 6.12 & 0.21 & 16.13 & 11.76 & 0.59 & 2.19 & 0.497 \\
\hyperref[ac]{MEF} & 17.20 & 0.69 & 74.06 & 6.13 & 0.26 & 14.70 & 11.23 & 0.56 & 2.83 & 0.499 \\
\hyperref[ac]{NPE} & 16.28 & 0.77 & 63.75 & 7.10 & 0.18 & 23.91 & 12.91 & 0.53 & 2.64 & 0.498 \\
\hyperref[ac]{VV} & 17.45 & 0.80 & 91.08 & 7.55 & 0.20 & 24.42 & 9.81 & 0.63 & 2.20 & 0.498 \\
\hyperref[ac]{SICE} (mix) & \textbf{\textcolor{Red}{15.04}} & \textbf{\textcolor{Red}{0.67}} & \textbf{\textcolor{Red}{125.23}} & 6.60 & \textbf{\textcolor{Red}{0.39}} & 21.88 & 11.51 & 0.54 & 3.02 & \textbf{\textcolor{Red}{0.496}} \\
\hyperref[ac]{SICE} (grad) & 15.45 & 0.72 & 80.86 & 6.32 & 0.35 & 21.98 & 11.16 & 0.54 & 2.95 & 0.497 \\
\bottomrule
\end{tabular}%
}
\label{tab:image_quality_results}
\end{table*}

\subsection{Architecture Overview}
\label{sec:arch}

RestoRect implements a two-stage knowledge distillation framework for efficient image restoration. Given degraded input $I_{LQ} \in \mathbb{R}^{H \times W \times 3}$ and ground truth $I_{GT} \in \mathbb{R}^{H \times W \times 3}$, the objective is:

$$\mathcal{F}_S(I_{LQ}) \approx \mathcal{F}_T(I_{LQ}) \approx I_{GT}$$

where $\mathcal{F}_T$ represents the teacher network (Stage 1) and $\mathcal{F}_S$ the student network (Stage 2).

\subsubsection{Retinex Decomposition Networks}

The Retinex decomposition models an image as the product of reflectance and illumination:
$$I = R \odot L$$

Two decomposition networks $\mathcal{D}_l$ (low-light) and $\mathcal{D}_h$ (normal-light) map:
$$\mathcal{D}(I) \rightarrow (R, L)$$

where $R \in \mathbb{R}^{H \times W \times 3}$ and $L \in \mathbb{R}^{H \times W \times 1}$.

\textbf{Network Architecture:}
\begin{align}
\text{Decom}(I) = &\text{ReLU}(\text{Conv2d}_{32 \rightarrow 4}^{3 \times 3}( \nonumber \\
&\text{LeakyReLU}_{0.2}(\text{Conv2d}_{32 \rightarrow 32}^{3 \times 3}( \nonumber \\
&\text{LeakyReLU}_{0.2}(\text{Conv2d}_{32 \rightarrow 32}^{3 \times 3}( \nonumber \\
&\text{LeakyReLU}_{0.2}(\text{Conv2d}_{3 \rightarrow 32}^{3 \times 3}(I))))))
\end{align}

Output split: $R = \text{output}[:, 0:3, :, :]$, $L = \text{output}[:, 3:4, :, :]$

\subsubsection{Feature Encoders}

\paragraph{Retinex ResNet Encoder (RRE)}

The RRE processes retinex features through separate reflectance and illumination pathways:

\textbf{Input Processing:}
$$\text{Retinex}_{LQ} = [R_{lq}; L_{lq}] \in \mathbb{R}^{H \times W \times 4}$$
$$\text{Retinex}_{GT} = [R_{gt}; L_{gt}] \in \mathbb{R}^{H \times W \times 4}$$

\textbf{Pixel Unshuffle:}
$$\text{X}_{0} = \text{PixelUnshuffle}_4(\text{Retinex}) \in \mathbb{R}^{H/4 \times W/4 \times 64}$$

\textbf{Channel Split:}
\begin{align}
X_R &= \text{X}_0[:, 0:48, :, :] \quad \text{(Reflectance channels)} \\
X_I &= \text{X}_0[:, 48:64, :, :] \quad \text{(Illumination channels)}
\end{align}

\textbf{Reflectance Branch ($E_R$):}
\begin{align}
E_R(X_R \oplus X_{R,gt}) = &\text{AdaptiveAvgPool2d}( \nonumber \\
&\text{LeakyReLU}_{0.1}(\text{Conv2d}_{128 \rightarrow 192}^{3 \times 3}( \nonumber \\
&\text{LeakyReLU}_{0.1}(\text{Conv2d}_{128 \rightarrow 128}^{3 \times 3}( \nonumber \\
&\text{LeakyReLU}_{0.1}(\text{Conv2d}_{64 \rightarrow 128}^{3 \times 3}( \nonumber \\
&\text{ResBlock}^6(\text{LeakyReLU}_{0.1}(\text{Conv2d}_{96 \rightarrow 64}^{3 \times 3}( \nonumber \\
&X_R \oplus X_{R,gt})))))))
\end{align}

\textbf{Illumination Branch ($E_I$):}
\begin{align}
E_I(X_I \oplus X_{I,gt}) = &\text{AdaptiveAvgPool2d}( \nonumber \\
&\text{LeakyReLU}_{0.1}(\text{Conv2d}_{128 \rightarrow 64}^{3 \times 3}( \nonumber \\
&\text{LeakyReLU}_{0.1}(\text{Conv2d}_{128 \rightarrow 128}^{3 \times 3}( \nonumber \\
&\text{LeakyReLU}_{0.1}(\text{Conv2d}_{64 \rightarrow 128}^{3 \times 3}( \nonumber \\
&\text{ResBlock}^6(\text{LeakyReLU}_{0.1}(\text{Conv2d}_{32 \rightarrow 64}^{3 \times 3}( \nonumber \\
&X_I \oplus X_{I,gt})))))))
\end{align}

\textbf{Feature Fusion:}
\begin{align}
\text{feat}_R &= \text{MLP}_R(E_R(\text{output})) \in \mathbb{R}^{192} \\
\text{feat}_I &= \text{MLP}_I(E_I(\text{output})) \in \mathbb{R}^{64} \\
\text{IPR}_{rex} &= [\text{feat}_R; \text{feat}_I] \in \mathbb{R}^{256}
\end{align}

\paragraph{Image ResNet Encoder (IRE)}

The IRE processes raw image features:

\textbf{Input Processing:}
\begin{align}
\text{X}_{LQ} &= \text{PixelUnshuffle}_4(I_{LQ}) \in \mathbb{R}^{H/4 \times W/4 \times 48} \\
\text{X}_{GT} &= \text{PixelUnshuffle}_4(I_{GT}) \in \mathbb{R}^{H/4 \times W/4 \times 48} \\
\text{X}_{concat} &= [\text{X}_{LQ}; \text{X}_{GT}] \in \mathbb{R}^{H/4 \times W/4 \times 96}
\end{align}

\textbf{Encoder Architecture:}
\begin{align}
E(\text{X}_{concat}) = &\text{AdaptiveAvgPool2d}( \nonumber \\
&\text{LeakyReLU}_{0.1}(\text{Conv2d}_{128 \rightarrow 256}^{3 \times 3}( \nonumber \\
&\text{LeakyReLU}_{0.1}(\text{Conv2d}_{128 \rightarrow 128}^{3 \times 3}( \nonumber \\
&\text{LeakyReLU}_{0.1}(\text{Conv2d}_{64 \rightarrow 128}^{3 \times 3}( \nonumber \\
&\text{ResBlock}^6(\text{LeakyReLU}_{0.1}(\text{Conv2d}_{96 \rightarrow 64}^{3 \times 3}( \nonumber \\
&\text{X}_{concat})))))))
\end{align}

\textbf{Output:}
$$\text{IPR}_{img} = \text{LayerNorm}(\text{MLP}(E(\text{output}))) \in \mathbb{R}^{256}$$

\subsubsection{UNet Transformer Architecture}

\paragraph{Spatial Channel Layer Normalization (SCLN)}

\hyperref[ac]{SCLN} captures global image statistics across spatial and channel dimensions:

\begin{align}
\mu_{global} &= \frac{1}{B \cdot C \cdot H \cdot W} \sum_{b,c,h,w} x_{b,c,h,w} \\
\sigma_{global}^2 &= \frac{1}{B \cdot C \cdot H \cdot W} \sum_{b,c,h,w} (x_{b,c,h,w} - \mu_{global})^2 \\
\text{SCLN}(x) &= \frac{x - \mu_{global}}{\sqrt{\sigma_{global}^2 + \epsilon}} \cdot \gamma
\end{align}

where $\gamma \in \mathbb{R}^C$ is learnable channel-wise scaling.

\paragraph{Retinex Attention}

The Retinex attention mechanism uses separate conditioning for reflectance and illumination components:

\textbf{Feature Conditioning:}
\begin{align}
k_{v_r} &= \text{Linear}(k_v[0:192]) \in \mathbb{R}^{3C/4 \times 1 \times 1} \\
k_{v_i} &= \text{Linear}(k_v[192:256]) \in \mathbb{R}^{C/4 \times 1 \times 1} \\
x_r &= x[:, 0:3C/4, :, :] \odot k_{v_r} + x[:, 0:3C/4, :, :] \\
x_i &= x[:, 3C/4:C, :, :] \odot k_{v_i} + x[:, 3C/4:C, :, :]
\end{align}

\textbf{Query-Key-Value Computation:}
\begin{align}
Q &= \text{DepthwiseConv}(\text{Conv}(x_r)) \in \mathbb{R}^{B \times C \times H \times W} \\
KV &= \text{DepthwiseConv}(\text{Conv}(x_i)) \in \mathbb{R}^{B \times 2C \times H \times W} \\
K, V &= \text{split}(KV, \text{dim}=1)
\end{align}

\textbf{Attention with \hyperref[ac]{QK} Normalization:}
\begin{align}
Q_{norm} &= \text{LayerNorm}(Q), \quad K_{norm} = \text{LayerNorm}(K) \\
Q_{norm} &= \frac{Q_{norm}}{\|Q_{norm}\|_2}, \quad K_{norm} = \frac{K_{norm}}{\|K_{norm}\|_2} \\
\text{Attn} &= \text{softmax}\left(\frac{Q_{norm} \cdot K_{norm}^T}{\sqrt{d_k}} \cdot \tau\right) \\
\text{Output} &= \text{Attn} \cdot V
\end{align}

where $\tau$ is a learnable temperature parameter.

\paragraph{Multi-Scale U-Net Architecture}

\textbf{Encoder Path:}
\begin{align}
\text{Level 1:} \quad &[B, 48, H, W] \xrightarrow{4 \times \text{TransformerBlock}} [B, 48, H, W] \\
&\downarrow \text{Downsample} \\
\text{Level 2:} \quad &[B, 96, H/2, W/2] \xrightarrow{6 \times \text{TransformerBlock}} [B, 96, H/2, W/2] \\
&\downarrow \text{Downsample} \\
\text{Level 3:} \quad &[B, 192, H/4, W/4] \xrightarrow{6 \times \text{TransformerBlock}} [B, 192, H/4, W/4] \\
&\downarrow \text{Downsample} \\
\text{Level 4:} \quad &[B, 384, H/8, W/8] \xrightarrow{8 \times \text{TransformerBlock}} [B, 384, H/8, W/8]
\end{align}

\textbf{Decoder Path with Skip Connections:}
\begin{align}
\text{Level 3:} \quad &\text{Upsample} + \text{Concat} + \text{ReduceChannel} \xrightarrow{6 \times \text{TransformerBlock}} \\
\text{Level 2:} \quad &\text{Upsample} + \text{Concat} + \text{ReduceChannel} \xrightarrow{6 \times \text{TransformerBlock}} \\
\text{Level 1:} \quad &\text{Upsample} + \text{Concat} \xrightarrow{4 \times \text{TransformerBlock}} \\
&\xrightarrow{4 \times \text{TransformerBlock}} \text{Conv2d}(96 \rightarrow 3) + \text{Residual}
\end{align}

\subsubsection{Auxiliary Constraints}

\paragraph{Anisotropic Diffusion}

The anisotropic diffusion operator preserves edges while smoothing noise:

$$\mathcal{A}(I) = \nabla \cdot (c(|\nabla I|) \nabla I)$$

with diffusion coefficient:
$$c(|\nabla I|) = \exp\left(-\frac{|\nabla I|^2}{s^2}\right)$$

where $s \in [0.01, 1.0]$ is a learnable sensitivity parameter.

\textbf{Texture Consistency Loss:}
$$L_{tex} = \|\mathcal{A}(I_{input}) - \mathcal{A}(R_{pred})\|_1$$

\textbf{Illumination Smoothness Loss:}
$$L_{lum} = \sum_{i,j} w_{i,j} \left(|\nabla_x L_{i,j}|^2 + |\nabla_y L_{i,j}|^2\right)$$

where $w_{i,j} = \exp(-|\nabla L_{i,j}|)$ provides gradient-aware weighting.

\paragraph{Polarized \hyperref[ac]{HVI} Color Space}

The polarized \hyperref[ac]{HVI} transformation eliminates red discontinuity:

\begin{align}
H_{polar} &= C_k \cdot S \cdot \cos(\pi H / 3) \\
V_{polar} &= C_k \cdot S \cdot \sin(\pi H / 3) \\
I_{polar} &= I_{max} = \max(R, G, B)
\end{align}

where the adaptive intensity collapse factor is:
$$C_k = k \cdot \sin(\pi I_{max} / 2) + \epsilon$$

with learnable parameter $k \in [0.1, 5.0]$.

\textbf{Polarized Color Loss:}
$$L_{col} = \|H_{polar}^{pred} - H_{polar}^{gt}\|_1 + \|V_{polar}^{pred} - V_{polar}^{gt}\|_1 + \|I_{polar}^{pred} - I_{polar}^{gt}\|_1$$

\subsubsection{Teacher Training Objective}

The complete teacher training loss combines:

$$L_{teach} = L_{rec} + L_{vgg} + L_{sty} + \lambda_{tex} L_{tex} + \lambda_{col} L_{col} + \lambda_{lum} L_{lum}$$

where:
\begin{align}
L_{rec} &= \|I_{pred} - I_{gt}\|_1 \quad \text{(pixel loss)} \\
L_{vgg} &= \sum_{l} \lambda_l \|\phi_l(I_{pred}) - \phi_l(I_{gt})\|_2^2 \quad \text{(perceptual loss)} \\
L_{sty} &= \sum_{l} \|G_l(\phi_l(I_{pred})) - G_l(\phi_l(I_{gt}))\|_F^2 \quad \text{(style loss)}
\end{align}

with $\lambda_{tex} = 0.05$, $\lambda_{col} = 0.05$, $\lambda_{lum} = 0.2$.

\subsection{Stage 2: Student Network Architecture}

\subsubsection{Rectified Flow Formulation}

Rectified flow models feature synthesis through straight-line interpolation:

$$\mathbf{x}_t = (1-t) \mathbf{z} + t \mathbf{f}_{teach}, \quad t \in [0,1]$$

where $\mathbf{z} \sim \mathcal{N}(0, I)$ is noise and $\mathbf{f}_{teach}$ are teacher features.

\textbf{Velocity Field:}
$$\mathbf{v}(\mathbf{x}_t, t) = \frac{d\mathbf{x}_t}{dt} = \mathbf{f}_{teach} - \mathbf{z}$$

\subsubsection{Velocity Prediction Networks}

\textbf{Architecture for both $\epsilon_\theta^{rex}$ and $\epsilon_\theta^{img}$:}
\begin{align}
\text{VelocityPredictor}(\mathbf{x}_t, t, \mathbf{c}) = &\text{ResMLP}^5( \nonumber \\
&\text{LeakyReLU}_{0.1}(\text{Linear}_{513 \rightarrow 256}( \nonumber \\
&[\mathbf{c}; t_{norm}; \mathbf{x}_t])))
\end{align}

where the input is $[\mathbf{c}; t; \mathbf{x}_t] \in \mathbb{R}^{513}$ with time normalization $t_{norm} = t / t_{max}$.

\textbf{Velocity Matching Loss:}
$$L_{vel} = \mathbb{E}_{t,\mathbf{z},\mathbf{f}_{teach}} \left[\|\epsilon_\theta(\mathbf{x}_t, t, \mathbf{c}) - \mathbf{v}(\mathbf{x}_t, t)\|_2^2\right]$$

\subsubsection{\hyperref[ac]{ODE} Integration for Inference}

During inference, the \hyperref[ac]{ODE} is solved using Euler's method:

$$\mathbf{x}_{t+\Delta t} = \mathbf{x}_t + \Delta t \cdot \epsilon_\theta(\mathbf{x}_t, t, \mathbf{c})$$

with adaptive step sizing $\Delta t = 1.0 / N_{steps}$ for $N_{steps} \in [1, 5]$.

\subsubsection{\hyperref[ac]{FLEX} Knowledge Distillation Loss}

\paragraph{Cross-Normalization}
\hyperref[ac]{FLEX} uses student statistics for normalizing both teacher and student features at each layer $l$:
\begin{align}
\mu_{stud}^l &= \frac{1}{H_l W_l} \sum_{h,w} \mathbf{f}_{stud}^{l,h,w} \\
\sigma_{stud}^l &= \sqrt{\frac{1}{H_l W_l} \sum_{h,w} (\mathbf{f}_{stud}^{l,h,w} - \mu_{stud}^l)^2 + \epsilon} \\
\mathbf{f}_{teach}^{l,norm} &= \frac{\mathbf{f}_{teach}^l - \mu_{stud}^l}{\sigma_{stud}^l} \\
\mathbf{f}_{stud}^{l,norm} &= \frac{\mathbf{f}_{stud}^l - \mu_{stud}^l}{\sigma_{stud}^l}
\end{align}

\paragraph{Percentile-Based Outlier Detection}
For each layer $l$ and channel $c$, we compute:
\begin{align}
\tau_p^{l,c} &= \text{Percentile}(|\mathbf{f}_{stud}^{l,c,norm}|, p) \\
M_{reliable}^{l,c,h,w} &= \mathbb{I}[|\mathbf{f}_{stud}^{l,c,norm,h,w}| \leq \tau_p^{l,c}]
\end{align}
where $p = 95\%$ is the outlier percentile threshold.

\paragraph{Resolution-Aware Weighting}
Dynamic resolution weighting prevents high-resolution features from dominating:
\begin{equation}
w_l^{res} = \max\left(\left(\frac{H_{base} W_{base}}{H_l W_l}\right)^{0.25}, 0.1\right)
\end{equation}
where $(H_{base}, W_{base}) = (64, 64)$ and $(H_l, W_l)$ is the spatial resolution at layer $l$.

\paragraph{Complete \hyperref[ac]{FLEX} Loss}
The final \hyperref[ac]{FLEX} loss combines masked feature matching with dual weighting:
\begin{equation}
L_{FLEX} = \sum_{l} w_l^{layer} \cdot w_l^{res} \cdot \frac{\sum_{c,h,w} M_{reliable}^{l,c,h,w} \cdot \|\mathbf{f}_{teach}^{l,c,norm,h,w} - \mathbf{f}_{stud}^{l,c,norm,h,w}\|^2}{\sum_{c,h,w} M_{reliable}^{l,c,h,w} + \epsilon}
\end{equation}
where $w_l^{layer}$ are predefined layer importance weights and the denominator normalizes by the number of reliable (non-outlier) elements.

\subsubsection{Trajectory Consistency Regularization}

\textbf{Smooth Transitions:}
$$L_{trans} = \sum_{i=1}^{N-1} \|\mathbf{f}_{pred}^{i+1} - \mathbf{f}_{pred}^{i}\|_2^2$$

\textbf{Target Alignment:}
$$L_{target} = \|\mathbf{f}_{pred}^{final} - \mathbf{f}_{teach}\|_2^2$$

\textbf{Semantic Consistency:}
$$L_{cons} = \sum_{i=1}^{N} \text{cos\_dist}(\mathbf{f}_{pred}^{i}, \mathbf{f}_{teach})$$

\textbf{Complete Trajectory Loss:}
$$L_{traj} = \alpha_{trans} L_{trans} + \alpha_{target} L_{target} + \alpha_{cons} L_{cons}$$

with $\alpha_{trans} = 0.1$, $\alpha_{target} = 0.5$, $\alpha_{cons} = 0.2$.

\subsubsection{Two-Phase Training Protocol}

\textbf{Phase 1: Velocity Learning}
$$L_{phase1} = L_{vel}^{rex} + L_{vel}^{img} + \lambda_{KD} L_{KD} + \lambda_{traj} L_{traj}$$

\textbf{Phase 2: Full Network Training}
$$L_{phase2} = L_{rec} + \lambda_{FLEX} L_{FLEX} + \lambda_{vel} (L_{vel}^{rex} + L_{vel}^{img})$$

with $\lambda_{FLEX} = 0.15$, $\lambda_{vel} = 0.05$.

\subsection{Implementation Details}

\subsubsection{Network Dimensions and Parameters}

\textbf{Stage 1 (Teacher):}
\begin{itemize}
\item RGFormer dimensions: $\text{dim} = 48$
\item Multi-head attention heads: $[1, 2, 4, 8]$
\item Transformer blocks per level: $[4, 6, 6, 8]$
\item FFN expansion factor: $2.66$
\end{itemize}

\textbf{Stage 2 (Student):}
\begin{itemize}
\item Velocity predictor features: $256$
\item Rectified flow timesteps: $4$
\item \hyperref[ac]{ODE} integration steps: $1-5$
\end{itemize}

\subsubsection{Training Hyperparameters}

\textbf{Stage 1:}
\begin{itemize}
\item Learning rate: $2 \times 10^{-4}$
\item Batch size: $16$
\item Training iterations: $500k$
\end{itemize}

\textbf{Stage 2:}
\begin{itemize}
\item Phase 1 learning rates: $lr_{rex} = lr_{img} = 2 \times 10^{-4}$
\item Phase 2 learning rate: $1 \times 10^{-4}$
\item Phase 1 iterations: $50k$
\item Phase 2 iterations: $200k$
\end{itemize}


\end{document}